\documentclass[letterpaper]{article} % DO NOT CHANGE THIS
\usepackage{aaai2027}  % DO NOT CHANGE THIS
\usepackage[hyphens]{url}  % DO NOT CHANGE THIS
\usepackage{xcolor}
\usepackage[
    colorlinks=true,
    urlcolor=blue
]{hyperref}
\usepackage{graphicx} % DO NOT CHANGE THIS
\usepackage{natbib}  % DO NOT CHANGE THIS AND DO NOT ADD ANY OPTIONS TO IT
\usepackage{caption} % DO NOT CHANGE THIS AND DO NOT ADD ANY OPTIONS TO IT
\usepackage{algorithm}
\usepackage{newfloat}
\usepackage{listings}
\DeclareCaptionStyle{ruled}{labelfont=normalfont,labelsep=colon,strut=off} % DO NOT CHANGE THIS
\floatstyle{ruled}
\newfloat{listing}{tb}{lst}{}
\floatname{listing}{Listing}

\usepackage[dvipsnames]{xcolor}
\usepackage{multirow}
\usepackage{amsmath}
\usepackage{amssymb}
\usepackage{algpseudocode}
\usepackage{amsfonts}
\usepackage{graphicx}
\usepackage{bm,color}
\usepackage{colortbl}
\usepackage{rotating}
\usepackage{tabularx}
\usepackage{array}
\usepackage{url}
\usepackage[percent]{overpic}
\usepackage{listings}
\usepackage{booktabs}

\newcommand{\ourmethod}{Gaussian-JEPA}

\usepackage{makecell} % Added to support \makecell

\definecolor{pltblue}{RGB}{174, 199, 232}
\definecolor{pltorange}{RGB}{255, 229, 204}
\definecolor{pltgreen}{RGB}{204, 229, 204}
\definecolor{pltred}{RGB}{229, 204, 204}
\definecolor{pltpurple}{RGB}{239, 218, 230}

\definecolor{tabblue}{HTML}{1f77b4}
\definecolor{taborange}{HTML}{ff7f0e}
\definecolor{tabgreen}{HTML}{2ca02c}
\definecolor{tabred}{HTML}{d62728}
\definecolor{tabpurple}{HTML}{9467bd}

\definecolor{cblue}{RGB}{173, 201, 233}
\definecolor{clblue}{RGB}{222, 234, 246}
\definecolor{corange}{RGB}{255, 152, 67}
\definecolor{lorgange}{RGB}{255, 221, 149}

\definecolor{light-gray}{gray}{0.5}
\definecolor{pretty-blue}{RGB}{0, 113, 188}
\definecolor{rowcolor}{gray}{.95} % soft gray
\definecolor{linecolor}{gray}{.895} % soft gray

\makeatletter
\newcommand{\compressalg}{
  \setlength{\@tempdima}{\algorithmicindent}
  \renewcommand{\algorithmicindent}{\@tempdima}%
  \setlength{\itemsep}{0pt}
  \setlength{\parskip}{0pt}
  \setlength{\parsep}{0pt}
}
\makeatother

\usepackage{etoolbox}
\makeatletter
\AfterEndEnvironment{algorithm}{\let\@algcomment\relax}
\AtEndEnvironment{algorithm}{\kern2pt\hrule\relax\vskip3pt\@algcomment}
\let\@algcomment\relax
\newcommand\algcomment[1]{\def\@algcomment{\footnotesize#1}}
\renewcommand\fs@ruled{\def\@fs@cfont{\bfseries}\let\@fs@capt\floatc@ruled
  \def\@fs@pre{\hrule height.8pt depth0pt \kern2pt}%
  \def\@fs@post{}%
  \def\@fs@mid{\kern2pt\hrule\kern2pt}%
  \let\@fs@iftopcapt\iftrue}
\makeatother
\definecolor{codeblue}{rgb}{0.25,0.5,0.5}
\definecolor{codekw}{rgb}{0.85, 0.18, 0.50}

\def\ie{\emph{i.e.}}

\title{Gaussian-JEPA: Joint-Embedding Predictive Learning for 3D Gaussian Splats}

\author{
Bin Ren\textsuperscript{\rm 1},
Qi Ma\textsuperscript{\rm 2},
Yue Li\textsuperscript{\rm 3},
Zongyan Han\textsuperscript{\rm 1},
Yidi Li\textsuperscript{\rm 4},
Yuqian Fu\textsuperscript{\rm 5},
\\
Rao Muhammad Anwer\textsuperscript{\rm 1},
Theo Gevers\textsuperscript{\rm 3},
Fahad Shahbaz Khan\textsuperscript{\rm 1},
Salman Khan\textsuperscript{\rm 1}
}

\affiliations{
\textsuperscript{\rm 1}
Mohamed bin Zayed University of Artificial Intelligence (MBZUAI), AE
\\
\textsuperscript{\rm 2}
ETH Z\"urich, CH
\\
\textsuperscript{\rm 3}
University of Amsterdam, NL
\\
\textsuperscript{\rm 4}
Taiyuan University of Technology, CN
\\
\textsuperscript{\rm 5}
King Abdullah University of Science and Technology (KAUST), SA
}

\begin{document}

\nocopyright
\maketitle

\begin{abstract}
3D Gaussian Splatting (3DGS) represents 3D content with anisotropic primitives that jointly encode geometry and appearance. Fixed-budget encoders consume sampled observations of Gaussian assets, so the same object may be observed through different primitive realizations. Existing self-supervised methods mainly reconstruct masked Gaussian attributes, tying supervision to one sampled realization and requiring an input-space decoder. Latent prediction offers an alternative, but its application to Gaussian tokens requires targets that accommodate coupled attributes and heterogeneous spatial support. We introduce \ourmethod{}, which predicts representations of held-out Gaussian token blocks from visible context. An online encoder processes the context, while a shared exponential-moving-average encoder supplies stop-gradient features for multi-scale targets. Complementary target projections and feature-space grounding provide latent supervision without reconstructing Gaussian attributes. We evaluate the features under Gaussian resampling, partial observations, and renderable shape completion, together with transfer to part segmentation and object classification. Compared with matched reconstruction pretraining, \ourmethod{} is more consistent across resampled inputs, retains more instance information under partial observations, and provides stronger frozen features for Gaussian completion. These results support latent prediction as an effective objective for reusable 3D Gaussian representations. Code is on the \href{https://amazingren.github.io/Gaussian-JEPA/}{project page}.
\end{abstract}

%-----------------------%
\section{Introduction}
\label{sec:intro}
%-----------------------%
3D Gaussian Splatting (3DGS)~\cite{kerbl20233d} represents 3D content with anisotropic Gaussian primitives. Each primitive jointly encodes position, covariance, opacity, and appearance. This differentiable representation supports efficient novel-view synthesis and applications including 3D content generation and controllable editing~\cite{tang2024dreamgaussian,chen2024gaussianeditor}. Gaussian primitives are becoming a native input for learned scene understanding~\cite{li2025scenesplat,li2026chorus,huang2026longsplat}. These developments motivate learning reusable features from Gaussian assets. We focus on object-level assets, which provide a controlled setting for studying the representation objective before extending it to scenes.

\begin{figure}[!t]
    \centering
    \includegraphics[width=\linewidth]{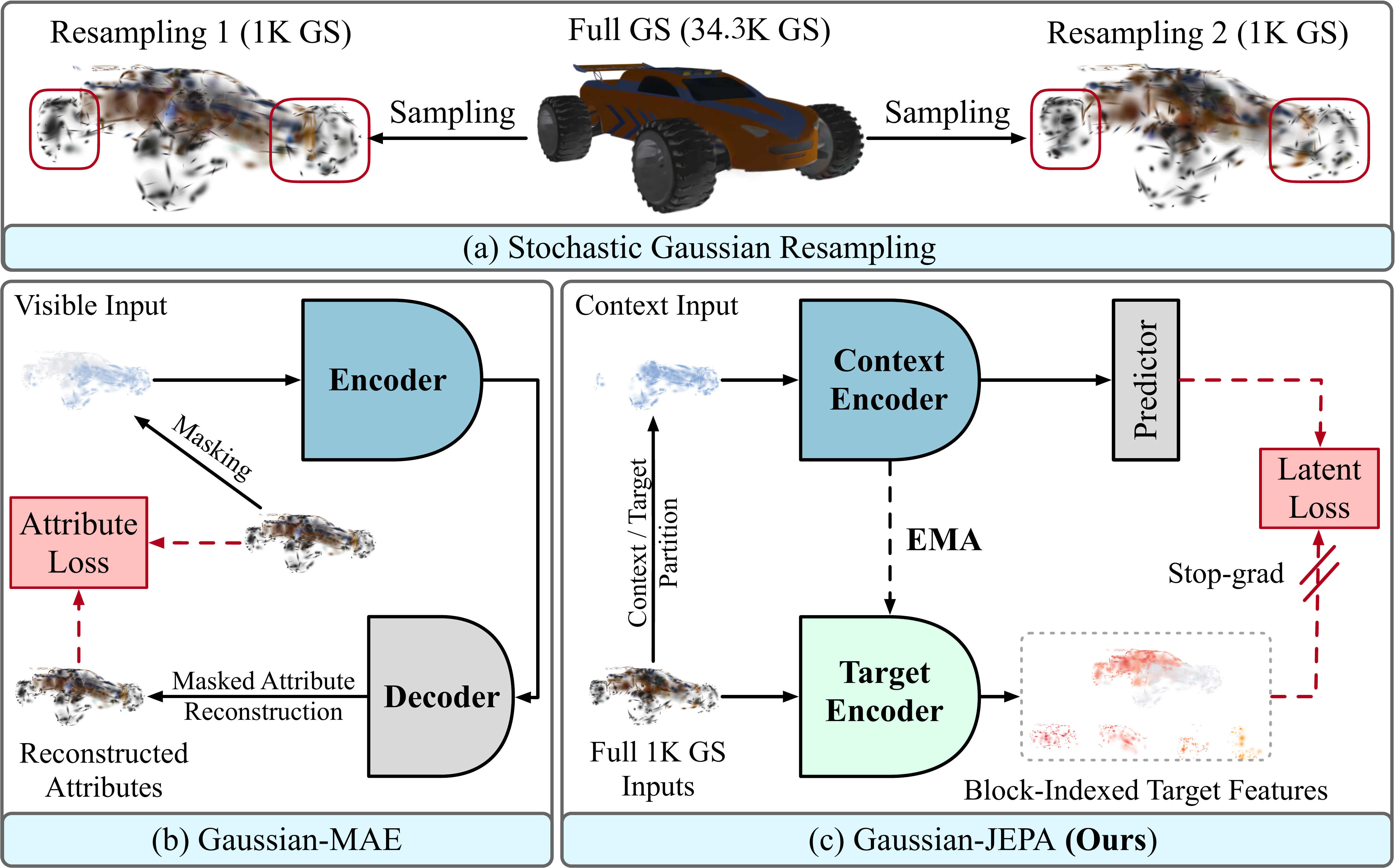}
    \caption{Motivation and learning paradigms.
    (a) Independent fixed-budget sampling yields different 1K GS observations
    of the same 34.3K GS asset.
    (b) Gaussian-MAE reconstructs masked attributes with an input-space decoder.
    (c) \ourmethod{} predicts block-indexed latent targets from visible context;
    the shared EMA encoder processes the full 1K GS input for each block. The
    lower panels use the same input and partition, isolating the learning objective.}
    \label{fig:teaser}
\end{figure}

Gaussian assets present a representation-specific difficulty. Practical encoders accept a fixed number of primitives, whereas an asset may contain many more. Independent fixed-budget resampling therefore produces different primitive observations for the same object, as shown in Fig.~\ref{fig:teaser}(a). The object remains unchanged, but its primitive-level observation does not. This variation arises at the interface between a dense asset and its encoder rather than from a designed semantic augmentation. Partial observations introduce a related challenge by removing coherent spatial regions. Existing self-supervised learning for Gaussian objects mainly follows masked autoencoding~\cite{devlin2019bert,he2022masked,pang2022masked}. Gaussian-MAE~\cite{ma2025large} reconstructs the centroids and attributes of masked local groups with a decoder, as shown in Fig.~\ref{fig:teaser}(b). This provides dense supervision, but ties its target to the primitives retained in a particular sample and requires decoding heterogeneous geometry and appearance attributes. Contrastive learning avoids reconstruction, but its learned invariances depend on the selected views and augmentations; global agreement also does not explicitly supervise missing spatial content. \textit{We therefore seek targets that capture predictable object structure rather than the exact attributes of one sampled realization.}

Joint-embedding predictive architectures (JEPA)~\cite{lecun2022path,assran2023self} offer a suitable principle: predict representations of hidden content from visible context without raw-input decoding or negative pairs. Point-JEPA~\cite{pointjepa2025} and 3D-JEPA~\cite{hu2024threedjepa} demonstrate this principle for point-cloud tokens. Its Gaussian instantiation is not immediate, however. A Gaussian token couples geometry, anisotropic support, visibility, and appearance; missing evidence may span different spatial extents; and latent targets must remain informative without attribute reconstruction. This setting raises three design questions: how to construct spatial targets, in which representation spaces to predict them, and how to regularize those spaces without a decoder.

We introduce \ourmethod{} to address these choices for object-level Gaussian assets. As illustrated in Fig.~\ref{fig:teaser}(c), local Gaussian tokens are divided into a shared context and non-overlapping target-token blocks at heterogeneous spatial scales. The online encoder processes the context once. For each block, a shared exponential-moving-average (EMA) encoder processes the complete token field, and the block indices select its contextualized target features. A lightweight predictor infers two complementary target projections from the context and target positions, while feature-space grounding regularizes their distributions. No raw Gaussian attribute appears in the prediction loss.

\textit{Our evaluation follows the representation problem rather than treating classification as the sole endpoint.} We measure frozen-feature consistency across independent resamplings, retrieve complete objects from partial queries, and train identical decoders to recover missing, renderable Gaussians from frozen features. Compared with matched Gaussian-MAE pretraining, \ourmethod{} produces more consistent representations across samples, retains more instance information under partial observations, and better supports Gaussian completion. Part segmentation and object classification provide complementary tests of semantic transfer.

Our contributions are threefold:
\begin{itemize}
    \item We formulate self-supervised object-level Gaussian learning under stochastic fixed-budget observations, replacing masked attribute reconstruction with prediction in representation space.
    \item We instantiate this formulation with heterogeneous multi-scale targets, complementary latent projections, and feature-space grounding, without an input-space decoder or raw-attribute loss.
    \item We evaluate Gaussian representations through resampling, partial observations, and renderable shape completion; part segmentation and object classification further assess semantic transfer.
\end{itemize}
%-----------------------%
\section{Related Work}
\label{sec:related}
%-----------------------%
\noindent \textbf{3D Gaussian Splatting and Gaussian pretraining.} 3D Gaussian Splatting~\cite{kerbl20233d,bao2025gaussian} represents a scene or object as anisotropic Gaussian primitives carrying centroid, opacity, scale, rotation, and view-dependent spherical harmonics. Its speed and fidelity have made 3DGS a widely used representation for reconstruction, editing, generation, and dynamic scenes. Gaussian-MAE~\cite{ma2025large} takes an important step toward reusable Gaussian features by constructing large-scale Gaussian object datasets and adapting masked autoencoding to 3DGS. Its pretext reconstructs raw Gaussian attributes with Gaussian feature grouping and splats pooling for heterogeneous attribute distributions. Recent 3DGS pretraining efforts further introduce semantics through rendered images or language supervision~\cite{you2026gaussfusion}. Our work studies a complementary, 3DGS-only direction based on latent prediction without raw-attribute reconstruction or external teachers.

\noindent \textbf{Self-supervised learning on point clouds.} Self-supervised 3D representation learning has largely followed reconstruction and contrastive paradigms. Point-BERT~\cite{PointBERT}, Point-MAE~\cite{pang2022masked}, and Point-M2AE~\cite{zhang2022point} adapt masked modeling to point patches, while methods such as CrossPoint~\cite{afham2022crosspoint} learn invariance across augmented views. MaskClu~\cite{ren2026masked} combines clustering with cluster-center reconstruction. Contrastive objectives can learn strong features, but their behavior depends on the selected transformations and positive/negative construction~\cite{ren2024bringing}. These methods primarily operate on point coordinates, optionally with normals or colors, whereas each 3D Gaussian jointly stores position, anisotropic support, opacity, orientation, and appearance. Our architecture and Gaussian-specific evaluations are designed for this richer primitive.

\noindent \textbf{Joint-embedding predictive learning.} JEPA~\cite{lecun2022path,assran2023self} predicts target representations from context without input reconstruction or negative pairs, echoing earlier predictability-maximization ideas~\cite{schmidhuber1993discovering}. Point-JEPA~\cite{pointjepa2025} orders adjacent point patches for latent prediction, while 3D-JEPA~\cite{hu2024threedjepa} predicts representations of spatial point-cloud blocks. CrossJEPA~\cite{perera2025crossjepa} instead transfers image supervision to a point-cloud encoder through cross-modal latent prediction. \ourmethod{} brings this paradigm to Gaussian primitives, couples geometry and appearance within each token, and grounds complementary target projections in feature space. Point-based JEPAs therefore provide broader context, while Gaussian-MAE remains the matched reconstruction baseline.

%-----------------------%
\section{Method}
\label{sec:method}
\begin{figure*}[t]
    \centering
    \includegraphics[width=\linewidth]{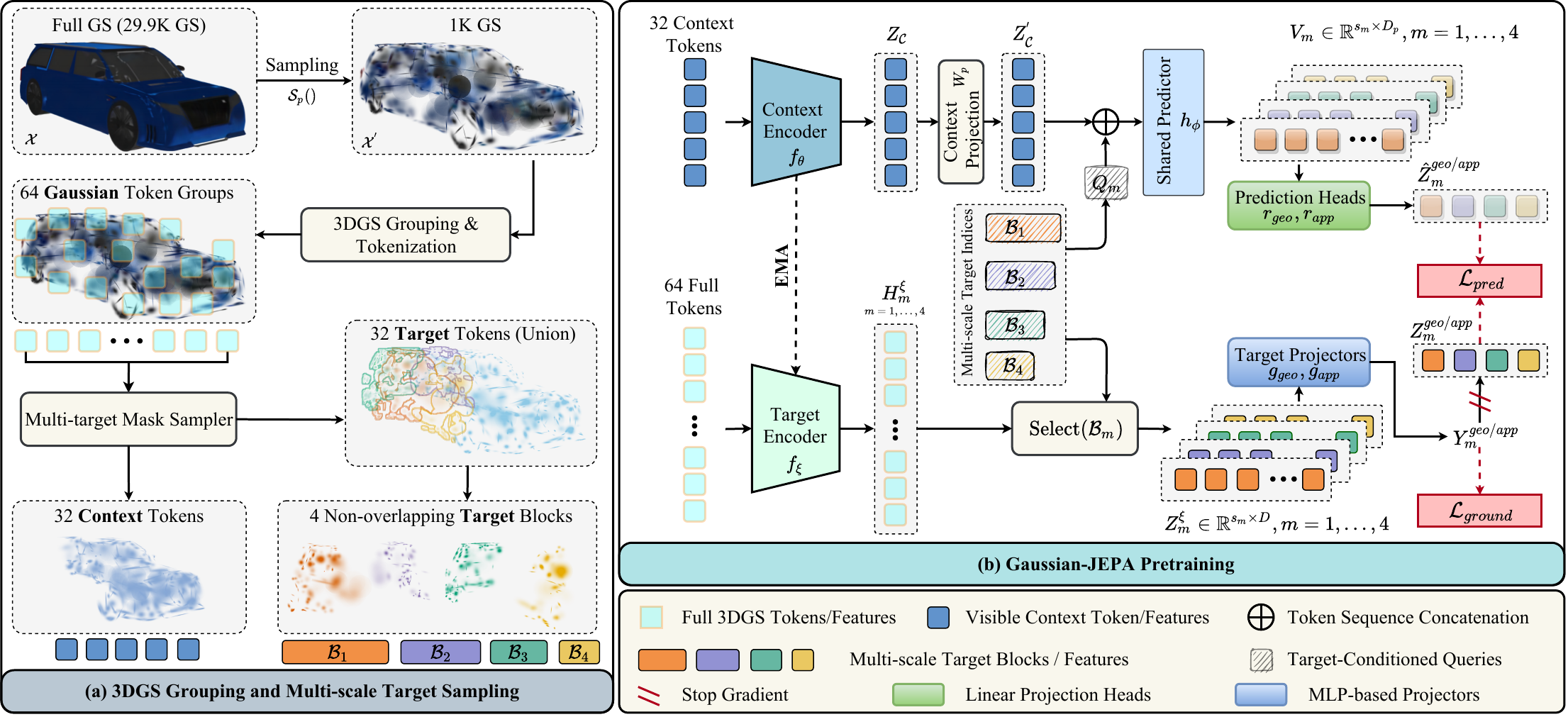}
    \caption{Overview of \ourmethod{}. 
    (a) A 1K-Gaussian observation is grouped into 64 local tokens and partitioned into a 32-token context and four non-overlapping target blocks. (b) The online encoder processes the shared context once. For each $\mathcal{B}_m$, the shared EMA encoder processes all tokens into $H_m^{\xi}$, whose indexed features form $Z_m^{\xi}$. A shared predictor combines context features with target-conditioned queries. Linear prediction heads and MLP target projectors define two complementary latent spaces. $\mathcal{L}_{\mathrm{pred}}$ aligns predicted and stop-gradient target features, while $\mathcal{L}_{\mathrm{ground}}$ regularizes target projections before detachment.}
    \label{fig:framework}
\end{figure*}

%-----------------------%
The proposed \ourmethod{} learns object-level 3D Gaussian representations through conditional prediction in latent space. Given a fixed-budget Gaussian observation, we form a shared visible context and several non-overlapping target-token blocks. The online encoder observes only the context, whereas a momentum encoder constructs each target from the complete token field. A shared predictor then infers the hidden block representations from context and target positions. Fig.~\ref{fig:framework} presents this information flow; Alg.~\ref{alg:gaussian_jepa} summarizes one training iteration.

%------------------------------------------------------------
\subsection{Problem Setup}
\label{sec:method:prelim}
%------------------------------------------------------------
\noindent\textbf{Gaussian object.} Following Gaussian-MAE~\cite{ma2025large}, an object is represented by a set of anisotropic Gaussian primitives,
\begin{equation}
    \mathcal{X}=[C,O,S,R,\mathrm{SH}]\in\mathbb{R}^{N\times14},
    \label{eq:gaussian_asset}
\end{equation}
where $C$, $O$, $S$, $R$, and $\mathrm{SH}$ denote centroids, opacity, scales, rotation quaternions, and DC spherical-harmonic coefficients. These parameters jointly describe primitive geometry, visibility, and appearance.

\noindent\textbf{Predictive formulation.} A Gaussian asset may contain more primitives than the encoder budget. We therefore observe it through a stochastic fixed-size sample $\mathcal{X}'\sim\mathcal{S}_{p}(\mathcal{X})$, where different draws may describe the same object with different primitive sets. Instead of reconstructing the attributes of one such realization, we partition its tokens into visible context $\mathcal{C}$ and hidden blocks $\{\mathcal{B}_m\}_{m=1}^{M}$, then predict the representation of each block from the shared context.

%------------------------------------------------------------
\subsection{Gaussian Tokens and Multi-Scale Targets}
\label{sec:method:tokens_targets}
%------------------------------------------------------------
\noindent\textbf{Gaussian tokenization.} Fig.~\ref{fig:framework}(a) starts from $p=1{,}024$ sampled Gaussians. We adopt the Gaussian-MAE tokenizer. Let $C'$ be the centroids in $\mathcal{X}'$. We select $n$ centers and gather $k$ neighbors for each center according to centroid distance:
\begin{equation}
\begin{aligned}
    U&=\operatorname{FPS}_{n}(C'),\\
    \mathcal{I}_i&=\operatorname{KNN}_{k}(\mathbf{u}_i,C'),\\
    P_i&=\mathcal{X}'[\mathcal{I}_i],\qquad
    \mathbf{t}_i=\psi(P_i),\\
    T&=[\mathbf{t}_1,\ldots,\mathbf{t}_n]
    \in\mathbb{R}^{n\times D}.
\end{aligned}
    \label{eq:gaussian_tokenization}
\end{equation}
KNN is determined by centroid coordinates, while each neighborhood retains 14 Gaussian attributes. Only its centroid is expressed relative to the group center before the PointNet-style tokenizer $\psi$. We use $n=64$, $k=32$, and $D=384$; an MLP $\rho$ embeds the centers $U$ as positional features. Further preprocessing and tokenization details are provided in the \textit{Supplementary Material} (\textit{Supp. Mat.}).

\noindent\textbf{Multi-scale target sampling.} Predicting equal-sized blocks exposes the model to a single target granularity. We instead specify a schedule $\mathbf{s}=[s_1,\ldots,s_M]$, with $s_1\geq\cdots\geq s_M$ and total target budget $n_t=\sum_m s_m$. Compact, non-overlapping blocks are sampled sequentially. Starting from $\mathcal{A}_0=\{1,\ldots,n\}$,
\begin{equation}
\begin{aligned}
    a_m&\sim\operatorname{Unif}(\mathcal{A}_{m-1}),\\
    \mathcal{B}_m&=\operatorname{NN}_{s_m}
    (\mathbf{u}_{a_m},U_{\mathcal{A}_{m-1}}),\\
    \mathcal{A}_m&=\mathcal{A}_{m-1}\setminus\mathcal{B}_m,
    \qquad \mathcal{C}=\mathcal{A}_M.
\end{aligned}
    \label{eq:multiscale_partition}
\end{equation}
Removing each selected block from the available set guarantees $\mathcal{B}_i\cap\mathcal{B}_j=\varnothing$ ($i\neq j$); the complement forms the shared context. A fixed $n_t$ separates target granularity from the mask ratio. Our default schedule is $M=4$ and $\mathbf{s}=[11,9,7,5]$, \ie, 32 of the 64 tokens are targets and the remaining 32 form $\mathcal{C}$, while individual targets cover different spatial supports.

%------------------------------------------------------------
\subsection{Joint-Embedding Prediction}
\label{sec:method:jepa}
%------------------------------------------------------------
\noindent\textbf{Asymmetric encoding.} Let $f_{\theta}$ and $f_{\xi}$ denote the online and EMA encoders. The online branch encodes the shared context once. For each target block, the shared EMA encoder processes the complete token field; the corresponding features are then selected by the block indices:
\begin{equation}
\begin{aligned}
    Z_{\mathcal{C}}
    &=f_{\theta}(T_{\mathcal{C}},\rho(U_{\mathcal{C}}))
    \in\mathbb{R}^{(n-n_t)\times D},\\
    H_m^{\xi}
    &=f_{\xi}(T,\rho(U))
    \in\mathbb{R}^{n\times D},\\
    Z_m^{\xi}
    &=H_m^{\xi}[\mathcal{B}_m]
    \in\mathbb{R}^{s_m\times D},
    \qquad m=1,\ldots,M.
\end{aligned}
    \label{eq:online_target_encoders}
\end{equation}
The parameters of $f_{\xi}$ are shared across all target passes. Encoding the complete field gives each selected target feature object-level context, while the online encoder receives neither target-token embeddings nor target-center indices as content inputs. Gradients do not enter $f_{\xi}$; after each optimizer step, its parameters follow the exponential moving average. The momentum schedule is specified in the \textit{Supp. Mat.}

\noindent\textbf{Target-conditioned prediction.} The context features are first mapped to predictor width $D_p$:
$Z'_{\mathcal{C}}=W_pZ_{\mathcal{C}}$. For $\mathcal{B}_m$, a learned query $\mathbf{q}\in\mathbb{R}^{D_p}$ is repeated once per target token, giving $Q_m=\mathbf{1}_{s_m}\mathbf{q}^{\mathsf T}$. Positional embeddings bind these queries to the selected target centers:
\begin{equation}
\begin{aligned}
    V_m=\Big[h_{\phi}\!\Big(
    &[Z'_{\mathcal{C}};Q_m],\\[-2pt]
    &[\rho_{\phi}(U_{\mathcal{C}});
      \rho_{\phi}(U_{\mathcal{B}_m})]
    \Big)\Big]_{\mathrm{qry}}
    \in\mathbb{R}^{s_m\times D_p},\\
    \widehat{Z}_m^b&=r_b(V_m)
    \in\mathbb{R}^{s_m\times D},
    \qquad b\in\{\mathrm{geo},\mathrm{app}\}.
\end{aligned}
    \label{eq:target_predictor}
\end{equation}
Only the query outputs are retained as $V_m$. The predictor receives the target positions but not their Gaussian-token content. All blocks share the context features, Transformer predictor $h_{\phi}$, and linear prediction heads $r_b$; their sizes change only the query-sequence length.

\noindent\textbf{Complementary target spaces.} Gaussian tokens jointly encode geometric and appearance-related factors. Two independently parameterized MLP projectors map the selected EMA features into complementary latent spaces:
\begin{equation}
\begin{aligned}
    Y_m^b&=g_b(Z_m^{\xi}),\\
    Z_m^b&=\operatorname{LN}\!\left(\operatorname{sg}(Y_m^b)\right),
    \qquad b\in\{\mathrm{geo},\mathrm{app}\}.
\end{aligned}
    \label{eq:complementary_targets}
\end{equation}
Here $Y_m^b,Z_m^b\in\mathbb{R}^{s_m\times D}$ and $\operatorname{sg}$ denotes stop-gradient. Both projectors consume features formed from all Gaussian attributes. Hence, $\mathrm{geo}$ and $\mathrm{app}$ name learned target spaces rather than a hard attribute split. The feature-space objective below regularizes each space while discouraging redundant linear factors between them.

%------------------------------------------------------------
\subsection{Latent-Space Learning Objectives}
\label{sec:method:grounding}
%------------------------------------------------------------
\noindent\textbf{Block-wise prediction.} At every scale, the predicted block is aligned with its position-matched, stop-gradient target in both latent spaces:
\begin{equation}
    \mathcal{L}_{\mathrm{pred}}
    =\frac{1}{M}\sum_{m=1}^{M}
    \sum_{b\in\{\mathrm{geo},\mathrm{app}\}}
    \operatorname{SmoothL1}(\widehat{Z}_m^b,Z_m^b).
    \label{eq:prediction_loss}
\end{equation}
This objective trains the online encoder, shared predictor, and prediction heads to predict target-block features without decoding Gaussian parameters.

\noindent\textbf{Feature-space grounding.} Because $Y_m^b$ is detached before forming the prediction target, $\mathcal{L}_{\mathrm{pred}}$ does not train the target projectors $g_b$. We instead regularize their pre-detach outputs entirely in feature space. For each branch, we concatenate tokens across the batch and all blocks,
$Y^b=[Y_1^b;\ldots;Y_M^b]\in\mathbb{R}^{L_t\times D}$, where $L_t=B_{\mathrm{batch}}\sum_m s_m$. Following VISReg~\cite{wu2026visreg}, $\mathcal{R}_{\mathrm{vis}}$ controls feature center and scale and matches random one-dimensional projections to a Gaussian reference. We further penalize cross-covariance between the two spaces:
\begin{equation}
\begin{aligned}
    \widetilde{Y}^b
    &=Y^b-\operatorname{mean}(Y^b),\\
    \Gamma
    &=\frac{1}{L_t}
    (\widetilde{Y}^{\mathrm{geo}})^{\mathsf T}
    \widetilde{Y}^{\mathrm{app}},\\
    \mathcal{L}_{\mathrm{ground}}
    &=\mathcal{R}_{\mathrm{vis}}(Y^{\mathrm{geo}})
    +\mathcal{R}_{\mathrm{vis}}(Y^{\mathrm{app}})
    +\frac{1}{D}\lVert\Gamma\rVert_F^2.
\end{aligned}
    \label{eq:grounding_loss}
\end{equation}
The branch-wise terms stabilize the individual target spaces, while the cross-covariance term discourages them from encoding the same linear factors. The complete construction of $\mathcal{R}_{\mathrm{vis}}$ is detailed in the \textit{Supp. Mat.}

\noindent\textbf{Overall objective.} The final objective is
\begin{equation}
    \mathcal{L}
    =\mathcal{L}_{\mathrm{pred}}
    +\lambda\mathcal{L}_{\mathrm{ground}},
    \qquad \lambda=0.1.
    \label{eq:total_loss}
\end{equation}
Feature-space grounding updates the target projectors, whereas the target encoder remains gradient-free and evolves only through EMA. All supervision is therefore applied to latent features: \ourmethod{} uses neither an input-space decoder nor reconstruction losses over centroids, opacity, scale, rotation, or appearance.

\begin{algorithm}[!t]
\caption{Gaussian-JEPA Pretraining}
\label{alg:gaussian_jepa}
\compressalg
\small
\begin{algorithmic}[1]
\Require Gaussian asset $\mathcal{X}$, target schedule $\mathbf{s}=[s_1,\ldots,s_M]$
\Require Encoders $f_{\theta},f_{\xi}$, predictor $h_{\phi}$, context map $W_p$
\Require Position maps $\rho,\rho_{\phi}$, query $\mathbf{q}$, projectors $g_b$, heads $r_b$
\Ensure Loss $\mathcal{L}$ and updated model parameters

\State $\mathcal{X}'\sim\mathcal{S}_{1024}(\mathcal{X})$; $(T,U)\leftarrow\operatorname{Tokenize}(\mathcal{X}')$
\State $(\{\mathcal{B}_m\}_{m=1}^{M},\mathcal{C})\leftarrow\operatorname{MultiScalePartition}(U,\mathbf{s})$
\State $Z_{\mathcal{C}}\leftarrow f_{\theta}(T_{\mathcal{C}},\rho(U_{\mathcal{C}}))$; $Z'_{\mathcal{C}}\leftarrow W_pZ_{\mathcal{C}}$
\State $\mathcal{L}_{\mathrm{pred}}\leftarrow0$
\For{$m=1,\ldots,M$}
    \State $H_m^{\xi}\leftarrow f_{\xi}(T,\rho(U))$; $Z_m^{\xi}\leftarrow H_m^{\xi}[\mathcal{B}_m]$ \Comment{no gradient}
    \State $Q_m\leftarrow\mathbf{1}_{s_m}\mathbf{q}^{\mathsf T}$
    \State $\Pi_m\leftarrow[\rho_{\phi}(U_{\mathcal{C}});\rho_{\phi}(U_{\mathcal{B}_m})]$
    \State $V_m\leftarrow\big[h_{\phi}([Z'_{\mathcal{C}};Q_m],\Pi_m)\big]_{\mathrm{qry}}$
    \For{$b\in\{\mathrm{geo},\mathrm{app}\}$}
        \State $Y_m^b\leftarrow g_b(Z_m^{\xi})$; $Z_m^b\leftarrow\operatorname{LN}(\operatorname{sg}(Y_m^b))$
        \State $\widehat{Z}_m^b\leftarrow r_b(V_m)$
        \State $\mathcal{L}_{\mathrm{pred}}\leftarrow\mathcal{L}_{\mathrm{pred}}+\operatorname{SmoothL1}(\widehat{Z}_m^b,Z_m^b)$
    \EndFor
\EndFor
\State $\mathcal{L}_{\mathrm{pred}}\leftarrow\mathcal{L}_{\mathrm{pred}}/M$
\State Concatenate $\{Y_m^b\}_m$ as $Y^b$; compute $\mathcal{L}_{\mathrm{ground}}$ by Eq.~\eqref{eq:grounding_loss}
\State $\mathcal{L}\leftarrow\mathcal{L}_{\mathrm{pred}}+\lambda\mathcal{L}_{\mathrm{ground}}$
\State Update trainable parameters; $\xi\leftarrow\tau\xi+(1-\tau)\theta$
\State \Return $\mathcal{L}$
\end{algorithmic}
\end{algorithm}

%-----------------------%
\section{Experiments}
\label{sec:exp}
%-----------------------%
We evaluate whether the learned features capture object content beyond a particular Gaussian sample and whether they transfer to semantic tasks. We study the former through resampling consistency, partial-observation robustness, and frozen-feature shape completion, and the latter through part segmentation and object classification. We then ablate the main design choices. Gaussian-MAE~\cite{ma2025large} serves as the matched reconstruction baseline under identical Gaussian inputs and evaluation protocols; point-based results under their original protocols provide broader context. Training and implementation details appear in the \textit{Supp. Mat.}.

\subsection{Gaussian Resampling Consistency}
\label{sec:exp:resampling}
\begin{figure}[!t]
    \centering
    \includegraphics[width=\linewidth]{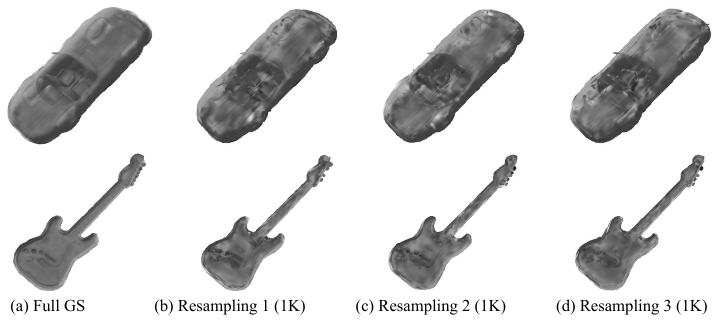}
    \caption{Gaussian resampling on ModelNet40-GS. Rows show \texttt{car\_0209} and \texttt{guitar\_0156}. (a) Full GS (15.7K/18.1K); (b)--(d) independent 1K resamplings.}
    \label{fig:resampling_inputs}
\end{figure}

\begin{figure}[t]
    \centering
    \includegraphics[width=\linewidth]{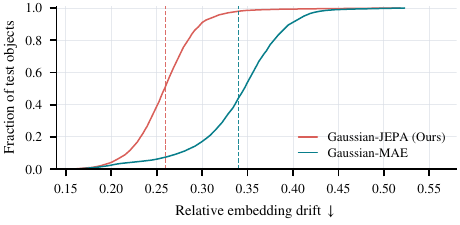}
    \caption{Relative embedding drift under Gaussian resampling. Each same-object distance is normalized by the mean distance to non-matching gallery objects in the same embedding space; dashed lines denote method means.}
    \label{fig:resampling_consistency}
\end{figure}
\begin{table}[!t]
    \centering
    {\small
    \setlength{\tabcolsep}{1mm}
    \begin{tabularx}{\columnwidth}{
        >{\raggedright\arraybackslash}Xccc
    }
    \toprule[0.81pt]
    Method
        & Rel. drift $\downarrow$
        & R@1 (\%) $\uparrow$
        & R@5 (\%) $\uparrow$ \\
    \midrule[0.6pt]
    Gaussian-MAE
        & 0.3400
        & 92.95
        & 98.89 \\
    \rowcolor{cyan!10}
    \ourmethod\ (\textit{Ours})
        & \textbf{0.2594}
        & \textbf{93.00}
        & \textbf{99.36} \\
    \bottomrule[0.81pt]
    \end{tabularx}
    }
    \caption{Frozen representation consistency under Gaussian resampling. Relative drift normalizes each same-object distance by its mean distance to all non-matching gallery objects in the corresponding embedding space. Retrieval uses one reference and four independentl resampled queries per ModelNet40-GS object.}
    \label{tab:resampling}
\end{table}

Gaussian assets often exceed an encoder's fixed input budget, so repeated sampling yields different primitive observations for the same object (Fig.~\ref{fig:resampling_inputs}). We evaluate frozen encoders on all 2,467 ModelNet40-GS test objects, unseen during pretraining. Five independent 1K samples are generated per object and shared by both methods; one forms the reference gallery and four form 9,868 queries. We concatenate mean- and max-pooled tokens and apply $\ell_2$ normalization.

Absolute distances from independently trained embedding spaces need not be calibrated identically. We therefore report relative drift
\begin{equation}
    d_{\mathrm{rel}}(i,q)=\frac{d(\mathbf{z}_{i}^{q},\mathbf{z}_{i}^{0})}{\frac{1}{N-1}\sum_{j\neq i}d(\mathbf{z}_{i}^{q},\mathbf{z}_{j}^{0})},
\end{equation}
which measures same-object sampling variation relative to each encoder's own inter-instance separation. Retrieval over the complete gallery supplies a complementary rank-based measure and guards against collapse.
As shown in Fig.~\ref{fig:resampling_consistency} and Tab.~\ref{tab:resampling}, \ourmethod{} reduces relative drift from 0.3400 to 0.2594, a relative reduction of $23.7\%$, and has lower object-averaged drift on $96.2\%$ of the test set. R@5 improves from $98.89\%$ to $99.36\%$, while R@1 remains comparable. Thus, its features are less sensitive to stochastic Gaussian sampling without sacrificing instance discrimination.

\subsection{Robustness to Partial Observations}
\label{sec:exp:partial_observation}
\begin{table}[t]
    \centering
    {\small
    \setlength{\tabcolsep}{1mm}
    \begin{tabularx}{\columnwidth}{
        >{\raggedright\arraybackslash}Xccc
    }
    \toprule[0.81pt]
    Method
        & R@1 (\%) $\uparrow$
        & R@5 (\%) $\uparrow$
        & R@10 (\%) $\uparrow$ \\
    \midrule[0.6pt]
    Gaussian-MAE
        & 19.80
        & 39.02
        & 49.14 \\
    \rowcolor{cyan!10}
    \ourmethod\ (\textit{Ours})
        & \textbf{39.82}
        & \textbf{65.28}
        & \textbf{75.95} \\
    \bottomrule[0.81pt]
    \end{tabularx}
    }
    \caption{Retrieval from partial Gaussian observations. At 55\% missing groups, five queries per object retrieve from the complete 2,467-object ModelNet40-GS gallery.}
    \label{tab:partial_observation}
\end{table}

\begin{figure}[t]
    \centering
    \includegraphics[width=\linewidth]{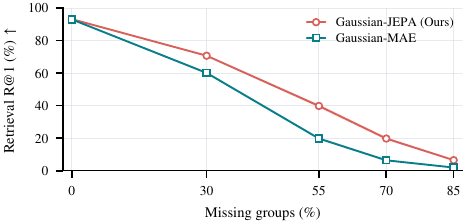}
    \caption{Robustness to partial Gaussian observations. Lines show five-seed mean R@1.
    % shaded regions show standard deviation.
    }
    \label{fig:partial_observation}
\end{figure}

Resampling changes the observed primitives while preserving object coverage; partial observation instead removes spatially coherent content. We retrieve complete ModelNet40-GS objects from cropped queries with missing ratios from $0\%$ to $85\%$. Both frozen encoders receive identical samples and masks, isolating their ability to preserve object identity as visible context decreases.
Fig.~\ref{fig:partial_observation} shows that \ourmethod{} degrades more slowly as spatial evidence disappears. At $55\%$ missing groups, Tab.~\ref{tab:partial_observation} reports gains of 20.02, 26.26, and 26.81 percentage points in R@1, R@5, and R@10. The R@1 advantage is already 10.56 points at $30\%$ missing and remains 13.36 points at $70\%$. This behavior indicates that latent prediction preserves more instance-level information under incomplete Gaussian observations.

\subsection{Gaussian Shape Completion}
\label{sec:exp:completion}
Recognition and retrieval test whether a representation preserves identity, whereas completion requires inferring missing geometry and appearance. On ShapeNet55-GS, a half-space crop forms a 512-Gaussian input and an independent 1K sample from the complete asset is the target. We freeze each encoder and train identical decoders without target coordinates, isolating representation quality. Evaluation uses held-out objects across three visibility ratios and crop seeds.

\begin{figure}[t]
    \centering
    \includegraphics[width=\linewidth]{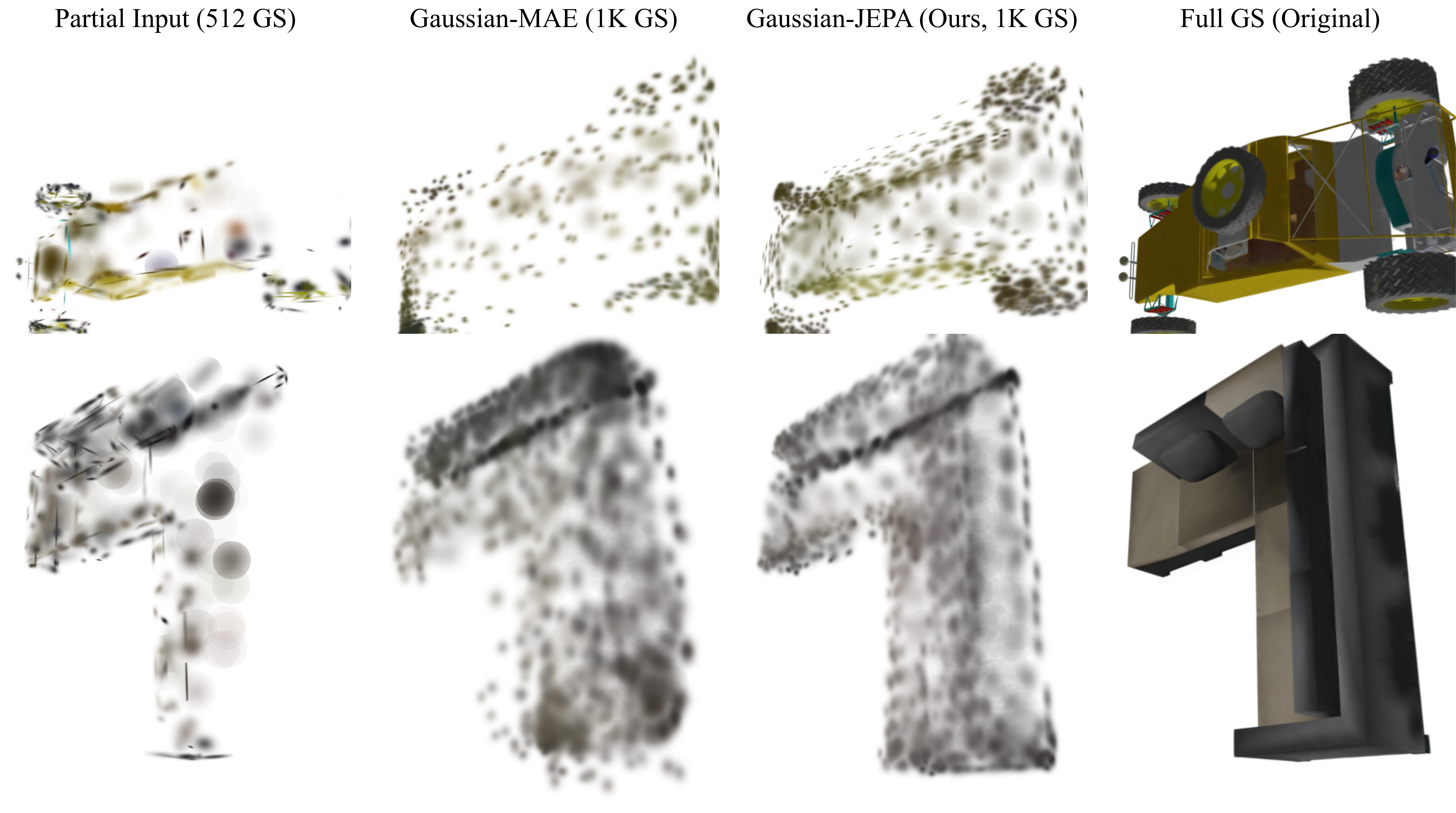}
    \caption{Qualitative Gaussian shape completion. Given a partial input of 512 Gaussians, decoders with identical architectures predict complete 1K-Gaussian representations from frozen Gaussian-MAE and Gaussian-JEPA features. The original full asset is shown only as a visual reference. Gaussian-JEPA recovers more coherent missing geometry.}
    \label{fig:shape_completion}
\end{figure}
\begin{table}[t]
    \centering
    {\small
    \setlength{\tabcolsep}{1.2pt}
    \begin{tabularx}{\columnwidth}{
        >{\raggedright\arraybackslash}Xccccc
    }
    \toprule[0.81pt]
    Method
        & CD $\downarrow$
        & F$_{1\%}$ $\uparrow$
        & PSNR $\uparrow$
        & PSNR$_{\mathrm{fg}}$ $\uparrow$
        & SSIM $\uparrow$ \\
    \midrule[0.6pt]
    Gaussian-MAE
        & 0.0732
        & 6.62
        & 16.03
        & 11.31
        & 0.7148 \\
    \rowcolor{cyan!10}
    \ourmethod\ (\textit{Ours})
        & \textbf{0.0678}
        & \textbf{7.42}
        & \textbf{17.24}
        & \textbf{12.38}
        & \textbf{0.7469} \\
    \bottomrule[0.81pt]
    \end{tabularx}
    }
    \caption{Shape completion on ShapeNet55-GS. CD and F$_{1\%}$ (\%) are averaged over three independently trained decoders; full statistics are provided in the \textit{Supp. Mat.} Render metrics use the seed-0 decoder at 50\% visibility; PSNR is in dB.}
    \label{tab:shape_completion}
\end{table}

Across three independently trained decoders, Tab.~\ref{tab:shape_completion} shows that \ourmethod{} reduces CD by $7.4\%$ and improves F$_{1\%}$ by 0.80 percentage points, with lower variation across seeds. The seed-0 render evaluation further improves PSNR by 1.21 dB, foreground PSNR by 1.07 dB, and SSIM by 0.0321. Together with Fig.~\ref{fig:shape_completion}, these results show that its frozen features better support the recovery of missing Gaussian geometry and renderable attributes.

\subsection{Part Segmentation}
\label{sec:exp:partseg}
\begin{table}[!t]
    \centering
    {\small
    \setlength{\tabcolsep}{3pt}
    \begin{tabularx}{\columnwidth}{
        >{\raggedright\arraybackslash}Xcc
    }
    \toprule[0.81pt]
    Method
        & mIoU$_C$ (\%) $\uparrow$
        & mIoU$_I$ (\%) $\uparrow$ \\
    \midrule[0.6pt]

    \rowcolor{rowcolor}
    \multicolumn{3}{c}{
        \textit{Supervised representation learning}
    } \\
    \midrule[0.6pt]

    PointNet~\cite{PointNet}
        & 80.4
        & 83.7 \\
    PointNet++~\cite{PointNet++}
        & 81.9
        & 85.1 \\
    Transformer~\cite{vaswani2017attention}
        & 83.4
        & 85.1 \\
    PTv1~\cite{zhao2021point}
        & 83.7
        & \textbf{86.6} \\

    \midrule[0.6pt]
    \rowcolor{rowcolor}
    \multicolumn{3}{c}{
        \textit{Self-supervised pretraining}
    } \\
    \midrule[0.6pt]

    Point-BERT$^{\dagger}$~\cite{PointBERT}
        & 84.1
        & 85.6 \\
    Point-MAE$^{\dagger}$~\cite{pang2022masked}
        & 84.2
        & \textbf{86.1} \\
    Point-JEPA$^{\dagger}$ (Saito et al. 2025)
        & 83.9
        & 85.8 \\
    Gaussian-MAE~\cite{ma2025large}
        & 84.2
        & 86.0 \\

    \rowcolor{cyan!10}
    \ourmethod\ (\textit{Ours})
        & \textbf{84.5} (+0.3)
        & \textbf{86.1} (+0.1) \\

    \bottomrule[0.81pt]
    \end{tabularx}
    }
    \caption{Part segmentation on ShapeNet-Part. $^{\dagger}$ denotes published point-cloud results under their original protocols. Both Gaussian methods use the same downstream setting; parenthesized values show gains over Gaussian-MAE.}
    \label{tab:partseg}
\end{table}
\begin{figure}[!t]
    \centering
    \includegraphics[width=\linewidth]{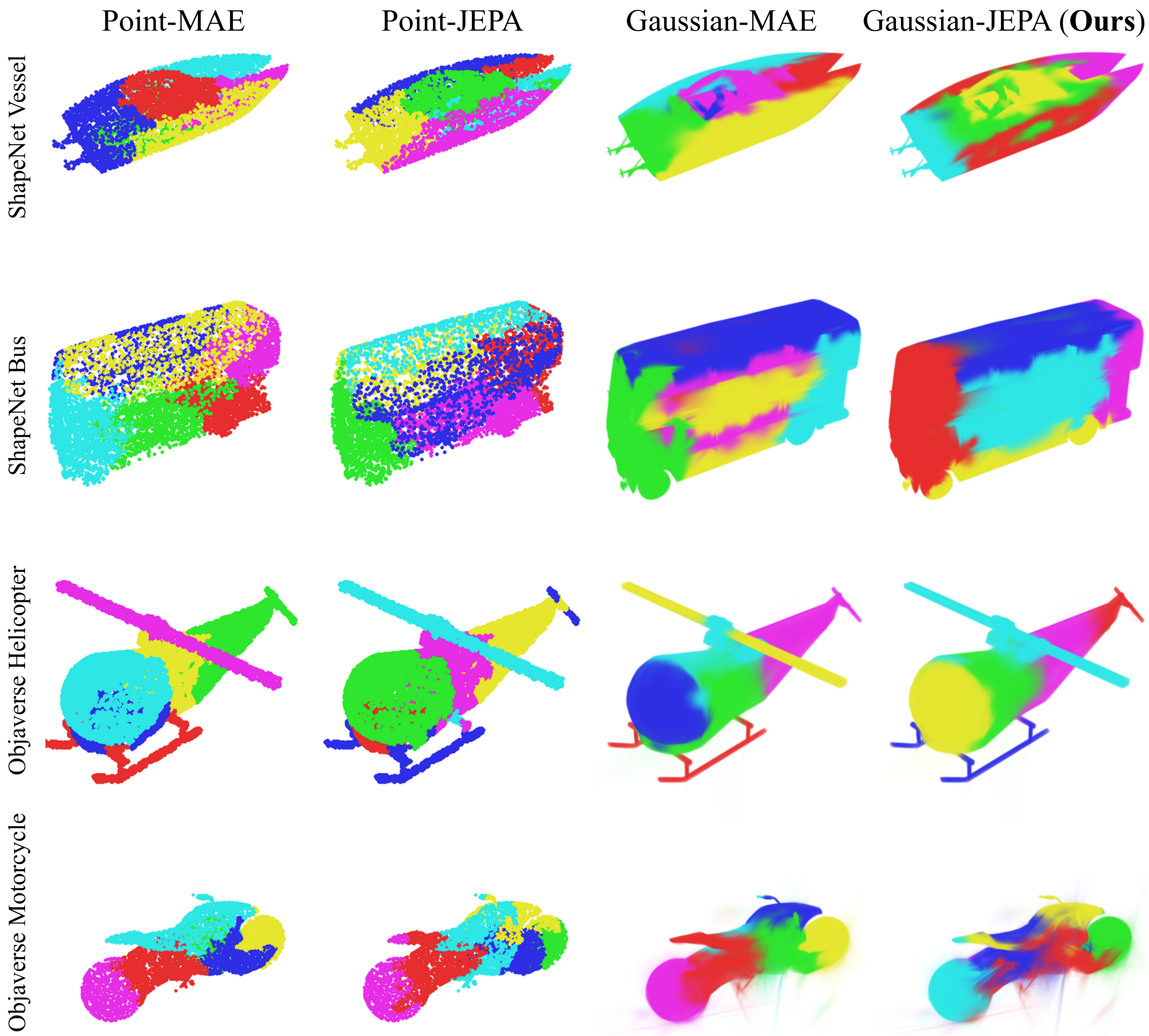}
    \caption{Clustering of encoded local features. We apply $k$-means ($k=6$) to features from Point-MAE, Point-JEPA, Gaussian-MAE, and \ourmethod{}. Each point or Gaussian is colored by its cluster assignment, providing a qualitative view of spatial organization in the learned local features.}
    \label{fig:kmeans_features}
\end{figure}
Part segmentation tests whether pretraining preserves local semantic structure. Following prior self-supervised point-cloud methods~\cite{PointBERT,pang2022masked,pointjepa2025}, we pretrain on ShapeNet and fine-tune on ShapeNet-Part. Published point-based methods are reported under their original protocols; Gaussian-MAE and \ourmethod{} use identical Gaussian attributes and downstream settings.
Tab.~\ref{tab:partseg} shows that \ourmethod{} reaches 84.5\% class mIoU and 86.1\% instance mIoU. It improves over Point-JEPA by 0.6/0.3 points and Gaussian-MAE by 0.3/0.1 points in class/instance mIoU, respectively. These results establish competitive transfer to dense semantic prediction alongside the Gaussian-specific evaluations above.
Fig.~\ref{fig:kmeans_features} complements the label-based evaluation by clustering encoded local features without semantic supervision. The resulting assignments provide a qualitative comparison of how the four representations organize local structure.

\subsection{Object Classification}
\label{sec:exp:classification}
\begin{table}[!t]
    \centering
    {\small
    \setlength{\tabcolsep}{1.0pt}
    \begin{tabularx}{\columnwidth}{
        >{\raggedright\arraybackslash}Xccc
    }
    \toprule[0.81pt]
    Method & Input & MN10 & MN40 \\
    \midrule[0.6pt]

    \rowcolor{rowcolor}
    \multicolumn{4}{c}{\textit{Supervised Learning Only}} \\
    \midrule[0.6pt]

    PointNet$^{\dagger}$~\cite{PointNet}
        & 1K P & -- & 89.2 \\
    PointNet++$^{\dagger}$~\cite{PointNet++}
        & 1K P & -- & 91.9 \\
    PTv1$^{\dagger}$~\cite{zhao2021point}
        & 1K P & -- & 90.6 \\
    PTv2$^{\dagger}$~\cite{wu2022point}
        & 1K P & -- & 91.6 \\

    \midrule[0.6pt]
    \rowcolor{rowcolor}
    \multicolumn{4}{c}{
        \textit{Full fine-tuning} ({\scshape Full})
    } \\
    \midrule[0.6pt]

    Point-BERT$^{\dagger}$~\cite{PointBERT}
        & 1K P & 94.82 & 93.20 \\
    Point-MAE$^{\dagger}$~\cite{pang2022masked}
        & 1K P & 94.93 & 93.20 \\
    Point-JEPA$^{\dagger}$ (Saito et al. 2025)
        & 1K P & -- & \textbf{93.80} \\
    Gaussian-MAE
        & 1K G & 94.16 & 92.54 \\
    \rowcolor{cyan!10}
    \ourmethod\ (\textit{Ours})
        & 1K G
        & \textbf{94.94} (+0.78)
        & 92.63 (+0.09) \\

    \midrule[0.6pt]
    \rowcolor{rowcolor}
    \multicolumn{4}{c}{
        \textit{Linear probing} ({\scshape Mlp-Linear})
    } \\
    \midrule[0.6pt]

    Point-BERT$^{\dagger}$~\cite{PointBERT}
        & 1K P & 93.06 & \textbf{90.56} \\
    Point-MAE$^{\dagger}$~\cite{pang2022masked}
        & 1K P & 93.17 & 90.24 \\
    Gaussian-MAE
        & 1K G & 93.50 & 88.97 \\
    \rowcolor{cyan!10}
    \ourmethod\ (\textit{Ours})
        & 1K G
        & \textbf{93.72} (+0.22)
        & 90.47 (+1.50) \\

    \midrule[0.6pt]
    \rowcolor{rowcolor}
    \multicolumn{4}{c}{
        \textit{Non-linear probing} ({\scshape Mlp-$3$})
    } \\
    \midrule[0.6pt]

    Point-BERT$^{\dagger}$~\cite{PointBERT}
        & 1K P & \textbf{94.27} & 91.82 \\
    Point-MAE$^{\dagger}$~\cite{pang2022masked}
        & 1K P & 93.61 & \textbf{92.63} \\
    Gaussian-MAE
        & 1K G & 93.39 & 87.72 \\
    \rowcolor{cyan!10}
    \ourmethod\ (\textit{Ours})
        & 1K G
        & 94.16 (+0.77)
        & 90.27 (+2.55) \\

    \bottomrule[0.81pt]
    \end{tabularx}
    }
    \caption{Classification on ModelNet (accuracy, \%). \texttt{P}/\texttt{G}: points/Gaussians. $^{\dagger}$ denotes published point-cloud results. Gaussian methods use 1K-Gaussian pretraining and transfer; parenthesized values show gains over Gaussian-MAE.}
    \label{tab:cls}
\end{table}
Object classification complements dense prediction with global semantic transfer. Following prior work~\cite{dong2022autoencoders,qi2023contrast}, we evaluate full fine-tuning, linear probing, and three-layer MLP probing. Gaussian-MAE and \ourmethod{} use the same 1K-Gaussian pretraining and downstream setting; point-cloud results are included under their original protocols.
Tab.~\ref{tab:cls} reports differences over Gaussian-MAE of 0.78/0.09 points with full fine-tuning, 0.22/1.50 points with linear probing, and 0.77/2.55 points with MLP-3 on ModelNet10/40. On MN40, the frozen-backbone gains are larger (1.50/2.55 points), where performance depends more on the pretrained representation.

\subsection{Ablation and Discussion}
\label{sec:exp:ablation}
\begin{table}[t]
    \centering
    {\small
    \setlength{\tabcolsep}{3pt}
    \begin{tabularx}{\columnwidth}{
        >{\raggedright\arraybackslash}Xcc
    }
    \toprule[0.81pt]
    Variant & MN10 LP $\uparrow$ & Resamp. R@1 $\uparrow$ \\
    \midrule[0.6pt]

    \rowcolor{rowcolor}
    \multicolumn{3}{c}{
        \textit{Target spaces and projector supervision}
    } \\
    Single + feature-space grounding
        & 93.06 & 90.85 \\
    Dual + attribute reconstruction
        & 93.38 & 88.18 \\
    Dual + feature-space grounding
        & \textbf{93.50} & \textbf{93.23} \\

    \midrule[0.6pt]
    \rowcolor{rowcolor}
    \multicolumn{3}{c}{
        \textit{Target count; 32 masked groups}
    } \\
    $M=2$: $[16,16]$
        & 93.17 & 92.17 \\
    $M=4$: $[8,8,8,8]$
        & 93.50 & \textbf{93.23} \\
    $M=6$: $[6,6,5,5,5,5]$
        & \textbf{93.83} & 91.89 \\

    \midrule[0.6pt]
    \rowcolor{rowcolor}
    \multicolumn{3}{c}{
        \textit{Scale spread; $M=4$, 32 masked groups}
    } \\
    $\delta=0$: $[8,8,8,8]$
        & 93.50 & \textbf{93.23} \\
    \rowcolor{cyan!10}
    $\delta=1$: $[11,9,7,5]$
        & \textbf{93.72} & 93.00 \\
    $\delta=2$: $[14,10,6,2]$
        & 92.95 & 92.63 \\

    \midrule[0.6pt]
    \rowcolor{rowcolor}
    \multicolumn{3}{c}{
        \textit{Input attributes; equal-scale targets}
    } \\
    Centroid + appearance
        & 92.62 & 87.21 \\
    Centroid + covariance
        & 93.17 & \textbf{94.14} \\
    All Gaussian attributes
        & \textbf{93.50} & 93.23 \\

    \bottomrule[0.81pt]
    \end{tabularx}
    }
    \caption{Ablations on latent targets, target sampling, and Gaussian attributes (in \%). MN10 LP measures semantic transfer; resampling R@1 measures retrieval across Gaussian resamples. Target-count and scale-spread
    comparisons retain 32 masked groups; selected settings are shaded.}
    \label{tab:ablation}
\end{table}
Unless noted, each ablation fixes the encoder and training schedule while varying one axis. ModelNet10 linear probing (MN10 LP) measures frozen semantic transfer, whereas cross-resampling R@1 tests whether instance identity survives independent Gaussian resampling. Reporting both reveals settings improving recognition without improving sampling stability. Selected configurations are shaded in Tab.~\ref{tab:ablation}.

\noindent\textbf{Why complementary target spaces?} With feature-space grounding fixed, a controlled single-space Gaussian-JEPA obtains 93.06\% MN10 LP and 90.85\% resampling R@1. Adding a second target projection raises them to 93.50\% and 93.23\%, respectively. Both projectors receive the same all-attribute EMA features: \emph{geometry} and \emph{appearance} denote learned latent views, not a manual split of Gaussian channels. The gains come from complementary target parameterizations rather than extra input information.

\noindent\textbf{Why feature-space grounding?} With dual projections, replacing raw-attribute reconstruction with feature-space grounding improves MN10 LP (93.38\% to 93.50\%) and resampling R@1 (88.18\% to 93.23\%), while removing the input-space decoder and per-attribute losses.

\noindent\textbf{How many target blocks?} We vary $M$ with a fixed 32-token target budget. Moving from two to four blocks improves both metrics. Six blocks yield the highest MN10 LP (93.83\%) but reduce R@1 to 91.89\%, whereas four blocks attain the strongest R@1 (93.23\%). We use four blocks as a balanced choice; increasing target count is not uniformly beneficial.

\noindent\textbf{Why heterogeneous target scales?} For $M=4$, we parameterize block sizes as $[8+3\delta,8+\delta,8-\delta,8-3\delta]$, preserving the same 32-token budget. Moderate variation ($\delta=1$) improves MN10 LP by 0.22 points over equal blocks, with a 0.23-point R@1 reduction. A larger spread ($\delta=2$) degrades both metrics relative to $\delta=0$. The selected $[11,9,7,5]$ schedule thus introduces useful scale diversity without excessively unbalancing the targets.

\noindent\textbf{What do native Gaussian attributes contribute?} With equal-size targets, centroid plus covariance outperforms centroid plus appearance on both metrics. All attributes yield the highest MN10 LP (93.50\%), whereas centroid plus covariance gives the strongest R@1 (94.14\%). Thus, appearance complements geometry for semantic transfer but does not uniformly improve resampling stability.

\noindent\textbf{How sensitive is the target budget?} The \textit{Supp. Mat.} varies the effective mask ratio from 37.5\% to 62.5\% while holding the number and relative scale of targets fixed. MN10 LP changes by only 0.22 points across this range. We therefore use 50\% as a balanced context--target allocation rather than claiming it as a task-specific optimum.

%-----------------------%
\section{Conclusion}
\label{sec:conclusion}
%-----------------------%
We introduced \ourmethod{}, \textit{a predictive self-supervised framework for object-level 3DGS representations}. It replaces raw-attribute reconstruction with multi-scale prediction in complementary latent spaces, avoiding an input-space decoder. The learned features are more consistent under resampling, more robust to partial observations, and stronger for frozen-feature shape completion than matched reconstruction pretraining, while transferring competitively to part segmentation and object classification. These results position latent prediction as a practical alternative to reconstruction for learning reusable representations of Gaussian assets.

% \section*{Acknowledgments}
% This work ...

\bibliography{aaai2027}

\begin{center}
    {\LARGE\bfseries Supplementary Material\par}
\end{center}
\vspace{1em}

% \clearpage
% \section{Acknowledgments}
% We thank ...
% This work was partially supported by ...

\setcounter{section}{0}
\setcounter{figure}{0}    
\setcounter{table}{0}   
\setcounter{page}{1}
\setcounter{equation}{0}

% \definecolor{customblue}{rgb}{0.25, 0.41, 0.88} %
% {
% \setcounter{tocdepth}{2}   %
% \hypersetup{linkcolor=customblue}
% \tableofcontents
% }

\renewcommand{\thetable}{\Alph{table}}
\renewcommand{\thefigure}{\Alph{figure}}
\renewcommand{\thesection}{\Alph{section}}
\renewcommand{\theequation}{\Alph{equation}}

This supplement documents the datasets, implementation, evaluation protocols, and additional analyses for \ourmethod{}. Our formulation adapts joint-embedding predictive learning~\cite{lecun2022path,assran2023self} to object-level Gaussian tokens. The presentation follows the information flow of the main framework: Gaussian preprocessing and tokenization, multi-scale target construction, latent prediction, and optimization. Unless stated otherwise, comparisons with Gaussian-MAE~\cite{ma2025large} use identical Gaussian inputs and evaluation protocols.

\section{Datasets and Splits}
\label{sec:supp:datasets}

\noindent\textbf{ShapeNet55-GS.}
We use the Gaussian assets released with Gaussian-MAE~\cite{ma2025large}, derived from ShapeNet objects~\cite{chang2015shapenet}. The provided split contains 51,934 training objects and 520 held-out test objects. Representation pretraining and shape-completion decoder training use only the training split; completion is evaluated on the 520 test objects.

\noindent\textbf{ModelNet10/40-GS.}
The Gaussian versions of ModelNet10 and ModelNet40 use the assets and object identities provided by Gaussian-MAE, derived from ModelNet~\cite{modelnet40,ma2025large}. We retain the standard splits: 3,991/908 train/test objects for ModelNet10 and 9,842/2,467 for ModelNet40. Both benchmarks are used for classification. The ModelNet40-GS test split also supports frozen resampling-consistency and partial-observation evaluation. No ModelNet object is used for pretraining.

\noindent\textbf{ShapeNet-Part.}
Part segmentation uses the 16 object categories, 50 part labels, and official train/validation/test partition of ShapeNet-Part~\cite{yi2016scalable}. Training combines the train and validation splits; evaluation uses the test split. Each annotated point set is paired by object identity with its Gaussian asset. The pretrained encoder processes the Gaussian input, and its token features are propagated to the annotated query points for label prediction.

\noindent\textbf{ScanObjectNN.}
The centroid-only transfer study uses the OBJ\_BG, OBJ\_ONLY, and PB\_T50\_RS variants of ScanObjectNN~\cite{scanobjectnn}. These experiments restrict Gaussian pretraining to centroid coordinates and fine-tune on the original point-only inputs. We report them as cross-representation evidence rather than matched Gaussian evaluation.

\noindent\textbf{Objaverse.}
Objaverse~\cite{deitke2023objaverse} contributes only the additional PCA examples in Fig.~\ref{fig:supp:pca}; it is not used for pretraining, model selection, or quantitative evaluation.

\section{Implementation Details}
\label{sec:supp:implementation}

\subsection{Gaussian Representation and Input Processing}
\label{sec:supp:input}
Following Gaussian-MAE~\cite{ma2025large}, each primitive has 14 values: centroid (3), opacity (1), scale (3), rotation quaternion (4), and DC spherical-harmonic coefficients (3). Each normalized source asset is first sampled with replacement to form an 8,192-entry buffer. At every pretraining iteration, the runner selects $p=1{,}024$ distinct buffer positions without replacement. Source indices may repeat within the buffer, so the resulting input is a stochastic fixed-size multiset rather than a guaranteed unique subset of the asset.

We apply deterministic attribute preprocessing before sampling. Centroids are centered and divided by their maximum radius, placing the object inside the unit sphere. Stored opacity logits pass through a sigmoid and are mapped to $[-1,1]$. Log-scales are exponentiated, adjusted by the spatial normalization, centered, and normalized in their three-dimensional attribute space. Rotation quaternions are normalized to unit length and sign-canonicalized by their first component. DC coefficients are multiplied by the spherical-harmonic constant, clipped to $[-0.5,0.5]$, and scaled by $2/\sqrt{3}$.

\subsection{Gaussian Grouping and Tokenization}
\label{sec:supp:tokenizer}
Grouping depends only on centroid coordinates. Centroid-based FPS selects $n=64$ centers, and hard KNN gathers $k=32$ neighbors per center. We express neighborhood centroids relative to the group center while retaining opacity, scale, rotation, and appearance in the normalized object frame. Thus, heterogeneous attribute scales do not affect neighborhood selection, although token content still uses all 14 values. We use hard KNN rather than Gaussian-MAE's optional soft-neighbor aggregation.

The local tokenizer $\psi$ is a two-stage PointNet-style network. A shared pointwise mapping transforms each $14$-D primitive through $14\!\rightarrow\!128\!\rightarrow\!256$. Max pooling produces a group descriptor that is concatenated with each local feature. A second shared mapping $512\!\rightarrow\!512\!\rightarrow\!384$, followed by max pooling, yields one $384$-D token per group. Batch normalization and ReLU follow the first layer of each stage. A separate $3\!\rightarrow\!128\!\rightarrow\!384$ MLP embeds the group centroids.

\subsection{Multi-Scale Target Construction}
\label{sec:supp:targets}
Targets are sampled independently for each object and sequentially over the available group centers. For block $\mathcal{B}_m$, we draw an available anchor and select its $s_m$ nearest available centers. The selected indices are removed before constructing the next block, guaranteeing pairwise non-overlap. The large-to-small schedule $\mathbf{s}=[11,9,7,5]$ assigns 32 of the 64 tokens to four target granularities; the exact complement forms the shared 32-token context. The integer schedule fixes the total target budget, so redistributing $s_m$ does not change the effective mask ratio.

Non-overlap applies to token-center indices. Because KNN neighborhoods are formed independently around different centers, context and target tokens may share underlying Gaussian primitives; the construction does not impose primitive-level disjointness.

When explicit \texttt{target\_sizes} are supplied, they determine block cardinalities directly. The scalar \texttt{mask\_ratio} retained in the configuration does not round target sizes; the effective ratio is $32/64=0.5$.

\subsection{Joint-Embedding Prediction}
\label{sec:supp:prediction}
The block indices are fixed before either branch is evaluated. For a mini-batch of size $B$, the online encoder processes the shared context once. For each target block, the same EMA encoder independently processes the complete 64-token field, after which the corresponding target features are selected by index:
\begin{equation}
\begin{aligned}
    Z_{\mathcal C}
    &=f_{\theta}(T_{\mathcal C},\rho(U_{\mathcal C}))
      \in\mathbb{R}^{B\times32\times384},\\
    H_m^{\xi}
    &=f_{\xi}(T,\rho(U))
      \in\mathbb{R}^{B\times64\times384},\\
    Z_m^{\xi}
    &=H_m^{\xi}[\mathcal{B}_m]
      \in\mathbb{R}^{B\times s_m\times384}.
\end{aligned}
\label{eq:supp:joint_embedding_shapes}
\end{equation}
The target passes share EMA parameters but are executed separately. Encoding the complete field contextualizes the selected features with respect to the full object observation; only block-indexed features enter the target projectors and prediction loss. The large-to-small schedule supplies several spatial supports under a fixed total target budget.

The context projection $W_p$ maps $Z_{\mathcal C}$ from 384 to 192 dimensions. For target $m$, one learned 192-D query is repeated $s_m$ times and concatenated with the projected context along the token sequence. A separate $3\!\rightarrow\!128\!\rightarrow\!192$ MLP embeds context and target centroids for the predictor. The shared four-layer Transformer processes this sequence; only its final $s_m$ query outputs are retained. Two linear heads map them to the $384$-D target spaces. The predictor thus receives target positions and cardinality, but no target-token embeddings: positional features act as spatial addresses rather than target content. It is evaluated once per block because $s_m$ varies, while its weights and the encoded context are shared across blocks.

\subsection{Complementary Target Projections and Feature-Space Grounding}
\label{sec:supp:grounding}
For each blockwise target representation $Z_m^\xi$, two independent $384\!\rightarrow\!384\!\rightarrow\!384$ GELU MLPs produce $Y_m^{\mathrm{geo}}$ and $Y_m^{\mathrm{app}}$. Both receive target features encoded from all 14 Gaussian attributes. The branch names denote learned target views; neither branch is assigned explicit geometry or appearance channels.

Each projected target is detached before layer normalization, $Z_m^b=\operatorname{LN}(\operatorname{sg}(Y_m^b))$. The Smooth L1 prediction loss therefore updates the online encoder, context projection, learned query, predictor position embedding, shared predictor, and linear prediction heads, but not the EMA encoder or target projectors. The target projectors are instead optimized by $\mathcal{L}_{\mathrm{ground}}$ on the pre-detach features $Y_m^b$.

\begin{table*}[!t]
    \centering
    {\small
    \setlength{\tabcolsep}{6pt}
    \renewcommand{\arraystretch}{1.08}
    \begin{tabularx}{\textwidth}{@{}
        >{\raggedright\arraybackslash}X
        >{\raggedright\arraybackslash}X
        >{\raggedright\arraybackslash}X@{}}
    \toprule[0.81pt]
    \textbf{Component} & \textbf{Setting} & \textbf{Value} \\
    \midrule[0.6pt]
    \addlinespace[2pt]
    \rowcolor{rowcolor}\multicolumn{3}{@{}l@{}}{\textbf{Input and target construction}} \\
    Input & Gaussian budget / feature dimension & 1,024 / 14 (all attributes) \\
    Grouping & token groups / neighbors / soft KNN & 64 / 32 / No \\
    Target sampler & blocks / sizes / effective mask & 4 / $[11,9,7,5]$ / 50\% \\
    \addlinespace[3pt]
    \rowcolor{rowcolor}\multicolumn{3}{@{}l@{}}{\textbf{Network architecture}} \\
    Online / EMA encoders & depth / width / heads / drop path & 12 / 384 / 6 / 0.1 \\
    Predictor input & context projection / query dimension & $384\!\rightarrow\!192$ / 192-D \\
    Predictor & depth / width / heads & 4 / 192 / 6 \\
    Target projectors & architecture & $384\!\rightarrow\!384\!\rightarrow\!384$ \\
    Prediction heads & architecture & linear $192\!\rightarrow\!384$ \\
    \addlinespace[3pt]
    \rowcolor{rowcolor}\multicolumn{3}{@{}l@{}}{\textbf{Optimization}} \\
    Feature-space grounding & SWD projections / loss weight & 256 / 0.1 \\
    EMA momentum & start / end & 0.996 / 0.9999 \\
    Optimizer & type / learning rate / weight decay & AdamW / $10^{-3}$ / 0.05 \\
    Schedule & epochs / warm-up / total batch size & 300 / 10 / 256 \\
    \bottomrule[0.81pt]
    \end{tabularx}
    }
    \caption{Pretraining configuration. The integer target schedule fixes the effective mask ratio.}
    \label{tab:supp:pretrain}
\end{table*}

% \begin{table*}[t]
%     \centering
%     \small
%     \setlength{\tabcolsep}{5pt}
%     \begin{tabularx}{\textwidth}{@{}>{\raggedright\arraybackslash}Xccccc@{}}
%     \toprule[0.81pt]
%     Method & GPU & Encoder (M) & Trainable / total (M)
%         & Step (s) $\downarrow$ & Peak memory (GiB) $\downarrow$ \\
%     \midrule[0.6pt]
%     Gaussian-MAE~\cite{ma2025large} & H200 NVL & 21.83 & 29.14 / 29.14 & \textbf{0.860} & \textbf{8.05} \\
%     \rowcolor{cyan!10}[0pt][0pt]\ourmethod{} & H200 NVL & 21.83 & \textbf{24.44} / 46.27 & 1.048 & 11.96 \\
%     \bottomrule[0.81pt]
%     \end{tabularx}
%     \caption{Parameters and pretraining resources. Measurements use
%     FP32, batch size 256, 1K Gaussians, and one NVIDIA H200 NVL. Step time and
%     peak allocated memory cover forward, backward, AdamW, and the EMA update,
%     excluding data loading; timing averages 30 steps after 8 warm-up steps.
%     Encoder parameters are those retained for downstream evaluation.}
%     \label{tab:supp:resources}
% \end{table*}

\begin{table*}[t]
    \centering
    \small
    \setlength{\tabcolsep}{5pt}
    \begin{tabularx}{\textwidth}{
        >{\raggedright\arraybackslash}Xccccc
    }
    \toprule[0.81pt]
    Method & GPU & Encoder (M) & Trainable / total (M)
        & Step (s) $\downarrow$ & Peak memory (GiB) $\downarrow$ \\
    \midrule[0.6pt]
    Gaussian-MAE~\cite{ma2025large}
        & H200 NVL
        & 21.83
        & 29.14 / 29.14
        & \textbf{0.860}
        & \textbf{8.05} \\
    \rowcolor{cyan!10}
    \ourmethod{}
        & H200 NVL
        & 21.83
        & \textbf{24.44} / 46.27
        & 1.048
        & 11.96 \\
    \bottomrule[0.81pt]
    \end{tabularx}
    \caption{Parameters and pretraining resources. Measurements use
    FP32, batch size 256, 1K Gaussians, and one NVIDIA H200 NVL. Step time and
    peak allocated memory cover forward, backward, AdamW, and the EMA update,
    excluding data loading; timing averages 30 steps after 8 warm-up steps.
    Encoder parameters are those retained for downstream evaluation.}
    \label{tab:supp:resources}
\end{table*}

Following VISReg~\cite{wu2026visreg}, we regularize the first- and second-order statistics and projected distribution of each target space. For completeness, concatenate these features over the mini-batch and all blocks as $Y^b\in\mathbb{R}^{L_t\times D}$, where $L_t=B\sum_m s_m$. For branch $b\in\{\mathrm{geo},\mathrm{app}\}$, let
\begin{equation}
\begin{aligned}
    \mu^b &= \frac{1}{L_t}\sum_{i=1}^{L_t}Y_i^b,\\
    \overline{Y}^b &= Y^b-\mathbf{1}(\mu^b)^\mathsf{T},\\
    \sigma_j^b &= \frac{\|\overline{Y}_{:j}^b\|_2}{\sqrt{L_t}}+\epsilon,\\
    \widetilde{Y}_{:j}^b
    &= \frac{\overline{Y}_{:j}^b}
    {\operatorname{sg}(\sigma_j^b)}.
\end{aligned}
\label{eq:supp:grounding_stats}
\end{equation}
We use $\epsilon=10^{-6}$. The centering and scale terms are
\begin{equation}
\begin{aligned}
    \mathcal{R}_{\mathrm{center}}^b
    &=\frac{1}{D}\|\mu^b\|_2^2,\\
    \mathcal{R}_{\mathrm{scale}}^b
    &=\frac{1}{D}\sum_{j=1}^{D}(\sigma_j^b-1)^2.
\end{aligned}
\label{eq:supp:center_scale}
\end{equation}
For distributional shape, we sample $K=256$ unit-normalized directions $W=[w_1,\ldots,w_K]$ at each training step. Let $P_{:k}^b$ be the sorted values of $\widetilde{Y}^bw_k$, and let $q_i=\Phi^{-1}(i/(L_t+1))$ be standard-normal quantiles. The sliced-Wasserstein term is
\begin{equation}
\begin{aligned}
    \mathcal{R}_{\mathrm{shape}}^b
    &=\frac{1}{L_tK}\sum_{i=1}^{L_t}\sum_{k=1}^{K}
      (P_{ik}^b-q_i)^2,\\
    \mathcal{R}_{\mathrm{vis}}(Y^b)
    &=\mathcal{R}_{\mathrm{center}}^b
      +\mathcal{R}_{\mathrm{scale}}^b\\
    &\quad+\mathcal{R}_{\mathrm{shape}}^b.
\end{aligned}
\label{eq:supp:visreg}
\end{equation}
The cross-covariance term is computed from the centered branch matrices as $\| (\overline{Y}^{\mathrm{geo}})^\mathsf{T}\overline{Y}^{\mathrm{app}}/L_t\|_F^2/D$. The branch-wise terms regularize each projection, while cross-covariance discourages shared linear factors. This regularizer updates only the target projectors. The EMA encoder remains gradient-free and is updated from the online encoder after each optimizer step. Neither objective reconstructs Gaussian coordinates or attributes.

\subsection{Architecture and Optimization}
\label{sec:supp:configuration}
Tab.~\ref{tab:supp:pretrain} summarizes the final configuration. The online and EMA encoders have identical 12-layer Transformer architectures but separate parameters. The EMA encoder is initialized from the online encoder and updated after every optimizer step. Both remain in training mode during pretraining, so stochastic depth is active in the online and blockwise EMA passes; stochastic layers are disabled for evaluation. The context projection, learned query, shared predictor, prediction heads, and target projectors are trainable modules outside the EMA encoder.

\paragraph{Target-side computation.}
The online encoder processes the shared context once. The EMA encoder and predictor are evaluated once per target block. These calls share parameters and carry no gradient through the EMA pathway, so parameter count does not increase with $M$. The default $M=4$ setting therefore uses four gradient-free EMA forwards per mini-batch. Target-encoder cost scales with $M$ because each pass encodes the complete token field; predictor cost also depends on the block cardinalities $\{s_m\}_{m=1}^{M}$. Here, \emph{decoder-free} denotes the absence of an input-space decoder and raw-attribute reconstruction objective, not lower pretraining cost than Gaussian-MAE~\cite{ma2025large}.

During pretraining, centroids receive independent per-axis scaling in $[2/3,3/2]$ and translation in $[-0.2,0.2]$; Gaussian scales receive the same multiplicative scaling. These operations follow the shared Gaussian feature-augmentation pipeline. They transform encoder inputs and are not interpreted as an exact reparameterization of the rendered Gaussian field. AdamW uses a base learning rate of $10^{-3}$ and weight decay 0.05. The learning rate warms from $10^{-6}$ for 10 epochs, then follows cosine decay to $10^{-6}$ over 300 epochs. Weight decay is disabled for biases, one-dimensional parameters, and learned tokens. EMA momentum increases linearly from 0.996 to 0.9999 over optimizer steps.

\paragraph{Reproducibility environment.}
Experiments use Python 3.9.19, PyTorch 2.0.1, and CUDA 11.8 on NVIDIA H200 GPUs. Training entry points expose a random seed that initializes Python, NumPy, and PyTorch; data-loader workers receive deterministic worker-specific seeds. Evaluation seed counts and aggregation units are stated with the corresponding protocols below.

\begin{table*}[!t]
    \centering
    {\small
    \setlength{\tabcolsep}{5pt}
    \begin{tabularx}{\textwidth}{@{}>{\raggedright\arraybackslash}Xccc@{}}
    \toprule[0.81pt]
    Metric & Gaussian-MAE & \ourmethod{} & Paired improvement [95\% CI] \\
    \midrule[0.6pt]
    Resampling relative drift $\downarrow$
        & 0.3400 & \textbf{0.2594} & $0.0806\ [0.0789,\ 0.0821]$ reduction \\
    Partial R@1 at 55\% missing $\uparrow$
        & 19.80 & \textbf{39.82} & $20.02\ [18.89,\ 21.16]$ points \\
    Completion CD $\downarrow$
        & $0.0732{\pm}0.0026$ & $\mathbf{0.0678{\pm}0.0023}$
        & $0.00542\ [0.00516,\ 0.00567]$ reduction \\
    Completion F$_{1\%}$ $\uparrow$
        & $6.62{\pm}0.63$ & $\mathbf{7.42{\pm}0.44}$
        & $0.80\ [0.68,\ 0.91]$ points \\
    \bottomrule[0.81pt]
    \end{tabularx}
    }
    \caption{Paired object-level uncertainty for Gaussian-specific evaluations.
    Positive improvements favor \ourmethod{}. Retrieval and F-score values are
    percentages. Completion entries report mean $\pm$ standard deviation over
    three decoder seeds; confidence intervals use 10,000 paired object
    bootstrap replicates.}
    \label{tab:supp:paired}
\end{table*}

\paragraph{Parameters and pretraining cost.}
Tab.~\ref{tab:supp:resources} compares both objectives with the same model-step benchmark. The retained downstream encoders have identical size. \ourmethod{} has fewer trainable pretraining parameters because it omits an input-space decoder, whereas its total parameter count includes the gradient-free EMA encoder. The blockwise target pathway increases pretraining step time by 21.9\% and peak allocated memory by 48.6\%. This overhead is confined to pretraining; downstream evaluation retains only the online encoder.

\section{Evaluation Protocols and Additional Results}
\label{sec:supp:evaluation}
\paragraph{Baseline evaluation.}
For the Gaussian-specific experiments in the main paper, we evaluate the all-attribute Gaussian-MAE encoder with exactly the same resampling, masking, retrieval, and decoder protocols as \ourmethod{}. Because these tasks were not reported for Gaussian-MAE, the shared pipeline controls downstream implementation choices and isolates the pretrained encoders.

\subsection{Gaussian-Specific Evaluation Protocols}
\paragraph{Resampling consistency.}
We evaluate all 2,467 ModelNet40-GS test objects, none of which appears in ShapeNet55-GS pretraining. Each object yields five independent 1K-Gaussian inputs. The first forms the reference gallery; the remaining four produce 9,868 queries. Global embeddings concatenate mean- and max-pooled frozen tokens and are $\ell_2$ normalized. For each query, relative drift divides the distance to its matching reference by the mean distance to all 2,466 non-matching references in the same embedding space. This dimensionless ratio measures sampling variation relative to each encoder's own instance separation, avoiding direct comparison of raw distances across embedding spaces. We also report full-gallery retrieval ranks.

\paragraph{Partial observations.}
The gallery contains one independently sampled complete input per object. Query groups are ordered along a random spatial direction, and a contiguous cap is retained. Missing ratios of $0\%$, $30\%$, $55\%$, $70\%$, and $85\%$ retain 64, 45, 29, 19, and 10 groups. Five query seeds yield 12,335 queries per ratio. Both encoders receive identical samples and masks.

\paragraph{Shape completion.}
Both pretrained encoders are frozen and paired with identical decoders. Four Transformer layers of width 256 and eight heads predict 256 coarse primitives, which a fixed $2\!\times\!2$ folding grid expands to 1,024 Gaussians. The decoder receives no target coordinates and predicts all 14 attributes. Training combines squared-$\ell_2$ Chamfer distance with nearest-neighbor attribute matching. Opacity, scale, and DC coefficients use $\ell_1$ loss; rotation uses sign-invariant quaternion similarity. The aggregate attribute term has weight 0.1.

Each decoder is trained for 80 epochs with AdamW, learning rate $10^{-3}$, weight decay 0.05, five warm-up epochs, and batch size 64. We train three decoder seeds per frozen encoder. Evaluation covers all 520 held-out objects, visible ratios $\{0.3,0.5,0.7\}$, and three crop seeds. Render metrics use the seed-0 decoder at 50\% visibility and four fixed views per object.

\subsection{Paired Object-Level Uncertainty}
\label{sec:supp:paired}
The Gaussian-specific evaluations compare both encoders on identical held-out objects, samples, and spatial crops. We estimate uncertainty with 10,000 paired bootstrap replicates over objects. Repeated samples and crops are averaged within each object before resampling; completion additionally averages the three independently trained decoders. In Tab.~\ref{tab:supp:paired}, the 95\% confidence interval for every principal effect excludes zero. This paired analysis shows that the observed gains under resampling, severe partial observation, and completion are not attributable to test-object composition. The intervals quantify held-out object variation under the stated protocols, not variation across independently pretrained encoders.

\subsection{Standard Downstream Protocols}
In the matched ModelNet10/40 experiments, Gaussian-MAE and \ourmethod{} use 1,024 Gaussians for both pretraining and transfer, 64 groups of 32, and all 14 attributes. Full fine-tuning, linear probing, and three-layer MLP probing run for 300 epochs with AdamW, learning rate $10^{-3}$, weight decay 0.05, five warm-up epochs, and total batch size 256. Frozen probing updates only the classifier. ShapeNet-Part uses 2,048 Gaussians, 128 groups of 32, batch size 16, learning rate $10^{-4}$, and 300 epochs. Published point-cloud results retain their original protocols and are marked with $\dagger$ in the main paper; they are not treated as matched Gaussian comparisons.

\paragraph{ModelNet reproducibility and protocol matching.}
ModelNet fine-tuning can exhibit appreciable run-to-run variation. The Gaussian-MAE release notes this variability, and the official Point-JEPA~\cite{pointjepa2025} implementation reports particularly high variance on ModelNet40. We therefore identify result provenance and avoid interpreting small cross-environment differences as method gains. Gaussian-MAE reports its headline classification results using 1K-Gaussian pretraining and 4K-Gaussian transfer. Tab.~\ref{tab:supp:modelnet_protocols} lists that paper result alongside our released-code rerun. The principal comparison instead fixes both methods to 1K-Gaussian pretraining and transfer; protocol variants are not mixed within a comparison.
\begin{table}[t]
    \centering
    {\small
    \setlength{\tabcolsep}{6pt}
    \begin{tabularx}{\columnwidth}{@{}>{\raggedright\arraybackslash}Xcc@{}}
    \toprule[0.81pt]
    Method & MN10 & MN40 \\
    \midrule[0.6pt]
    \multicolumn{3}{@{}l}{\textit{4K-Gaussian transfer}} \\
    Gaussian-MAE (paper)~\cite{ma2025large}
        & 95.37 & 93.35 \\
    Gaussian-MAE (our rerun)
        & 95.31 & 92.46 \\
    \ourmethod\ (Ours)
        & 95.15 & 93.21 \\
    \midrule[0.6pt]
    \multicolumn{3}{@{}l}{\textit{1K-Gaussian transfer}} \\
    Gaussian-MAE
        & 94.16 & 92.54 \\
    \ourmethod\ (main setting)
        & \textbf{94.94} & \textbf{92.63} \\
    \bottomrule[0.81pt]
    \end{tabularx}
    }
    \caption{ModelNet full fine-tuning across Gaussian input budgets
    (accuracy, \%). All encoders are pretrained with 1K Gaussians. The
    published row is taken from Gaussian-MAE; our rerun uses its official code
    under the same 1K-to-4K protocol. The main comparison transfers both methods
    with 1K Gaussians.}
    \label{tab:supp:modelnet_protocols}
\end{table}

\subsection{Controlled Pretraining Baselines}
\label{sec:supp:controlled_baselines}

\noindent\textbf{Pretraining versus random initialization.}
To separate representation pretraining from backbone capacity, we train the same Gaussian classification backbone from random initialization under the full fine-tuning protocol. All architectural and downstream optimization settings remain unchanged. Tab.~\ref{tab:supp:scratch} shows gains of 1.33 points on MN10 and 4.43 points on MN40 from \ourmethod{} initialization, with the larger effect on the more diverse MN40 benchmark.

\begin{table}[t]
    \centering
    {\small
    \setlength{\tabcolsep}{4pt}
    \begin{tabularx}{\columnwidth}{@{}
        >{\raggedright\arraybackslash}Xccc@{}}
    \toprule[0.81pt]
    Encoder initialization & Transfer & MN10 & MN40 \\
    \midrule[0.6pt]
    Random initialization & Full & 93.61 & 88.20 \\
    \ourmethod{} pretraining & Full & \textbf{94.94} & \textbf{92.63} \\
    \bottomrule[0.81pt]
    \end{tabularx}
    }
    \caption{Effect of Gaussian pretraining (accuracy, \%). The backbone and
    full fine-tuning protocol are identical; only encoder initialization
    differs.}
    \label{tab:supp:scratch}
\end{table}

\noindent\textbf{Point-JEPA-style Gaussian adaptation.}
We also implement a diagnostic Gaussian adaptation of Point-JEPA~\cite{pointjepa2025} in the same codebase. It matches the Transformer scale, Gaussian grouping, ShapeNet55-GS split, and downstream recipes, while retaining centroid-only input $E(C)$, greedy spatial ordering, contiguous target spans, and one latent target space. It uses neither attribute reconstruction nor feature-space grounding. Tab.~\ref{tab:supp:pointjepa_style} reports ModelNet transfer under the same centroid-only input and 1K-Gaussian protocols. Across the three transfer settings, \ourmethod{} improves MN10 by 2.76--3.05 points and MN40 by 6.49--8.34 points. Under this controlled setup, the direct point-style adaptation does not match our Gaussian formulation. Tab.~\ref{tab:domain} separately compares centroid-only Gaussian-MAE and \ourmethod{} under point-only transfer; combining the two tables would conflate downstream modalities.

\begin{table}[!t]
    \centering
    {\small
    \setlength{\tabcolsep}{5pt}
    \begin{tabularx}{\columnwidth}{@{}
        >{\raggedright\arraybackslash}Xccc@{}}
    \toprule[0.81pt]
    Method & Transfer & MN10 & MN40 \\
    \midrule[0.6pt]
    Point-JEPA-style $E(C)$ & Full & 90.86 & 84.35 \\
    \ourmethod{} $E(C)$ & Full & \textbf{93.91} & \textbf{92.03} \\
    \midrule[0.3pt]
    Point-JEPA-style $E(C)$ & Linear & 89.87 & 82.29 \\
    \ourmethod{} $E(C)$ & Linear & \textbf{92.87} & \textbf{88.78} \\
    \midrule[0.3pt]
    Point-JEPA-style $E(C)$ & MLP-3 & 90.20 & 81.80 \\
    \ourmethod{} $E(C)$ & MLP-3 & \textbf{92.96} & \textbf{90.14} \\
    \bottomrule[0.81pt]
    \end{tabularx}
    }
    \caption{Centroid-only Gaussian transfer (accuracy, \%). Both models are
    pretrained for 300 epochs and evaluated with the same 1K-Gaussian
    downstream protocols.}
    \label{tab:supp:pointjepa_style}
\end{table}

\subsection{Centroid-Only Cross-Representation Transfer}
Tab.~\ref{tab:domain} isolates transfer from Gaussian pretraining to point-only inputs. Both models are pretrained with centroid-only Gaussian tokens, denoted $E(C)$, and fully fine-tuned with the same point-cloud protocol. \ourmethod{} improves ModelNet40, OBJ\_BG, and PB\_T50\_RS, whereas Gaussian-MAE is stronger on OBJ\_ONLY. The result provides evidence that the predictive objective can transfer beyond Gaussian attributes without implying universal superiority over point-specific methods.

\begin{table}[t]
    \centering
    {\small
    \setlength{\tabcolsep}{1pt}
    \begin{tabularx}{\columnwidth}{@{}>{\raggedright\arraybackslash}Xcccc@{}}
    \toprule[0.81pt]
    Method & MN40 & OBJ\_BG & OBJ\_ONLY & PB\_T50\_RS \\
    \midrule[0.6pt]
    Gaussian-MAE $E(C)$ & 90.88 & 84.24 & \textbf{88.64} & 82.15 \\
    \ourmethod\ $E(C)$ & \textbf{92.17} & \textbf{87.43} & 87.26 & \textbf{83.14} \\
    \bottomrule[0.81pt]
    \end{tabularx}
    }
    \caption{Centroid-only transfer (accuracy, \%). Gaussian-pretrained
    encoders are fully fine-tuned on ModelNet40 and three ScanObjectNN
    variants.}
    \label{tab:domain}
\end{table}

\subsection{Additional Qualitative Results}
\begin{figure}[t]
    \centering
    \includegraphics[width=\linewidth]{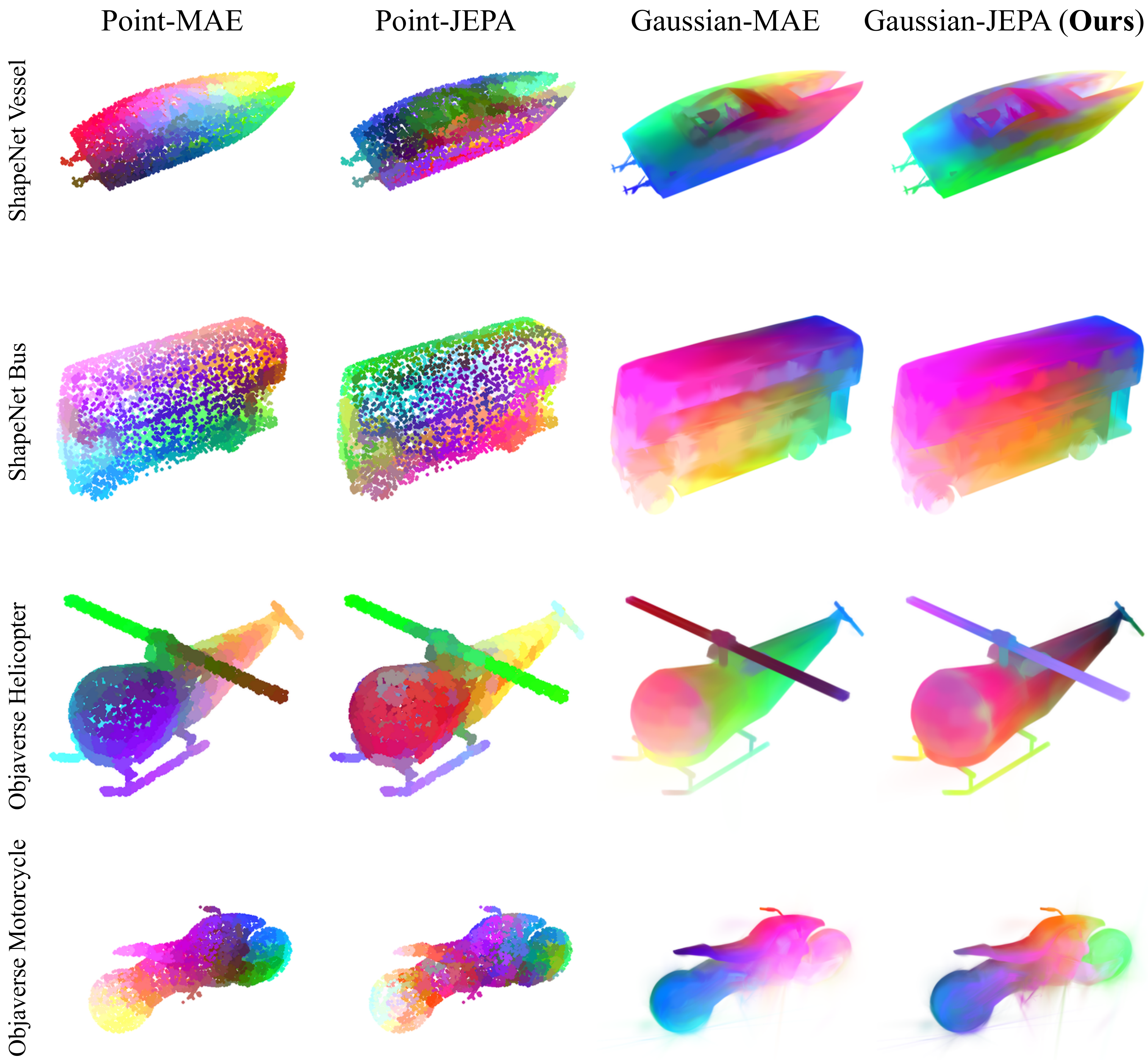}
    \caption{PCA visualization of frozen encoder features.
    Columns compare Point-MAE, Point-JEPA, Gaussian-MAE, and \ourmethod{};
    rows show ShapeNet and Objaverse objects. The first three principal
    components of propagated token features are mapped to RGB. Colors are
    interpreted within each panel and are not aligned across methods.}
    \label{fig:supp:pca}
\end{figure}

\noindent\textbf{PCA feature fields.}
Fig.~\ref{fig:supp:pca} complements the discrete K-means visualization in the main paper with continuous frozen feature fields. We propagate token features to the input primitives, project them onto the first three principal components, and map the components to RGB. PCA is fitted independently for each method--object pair, so absolute colors are not comparable across panels. In the displayed cases, \ourmethod{} yields coherent fields over extended structures while preserving variation between object parts.

\begin{figure}[t]
    \centering
    \includegraphics[width=\linewidth]{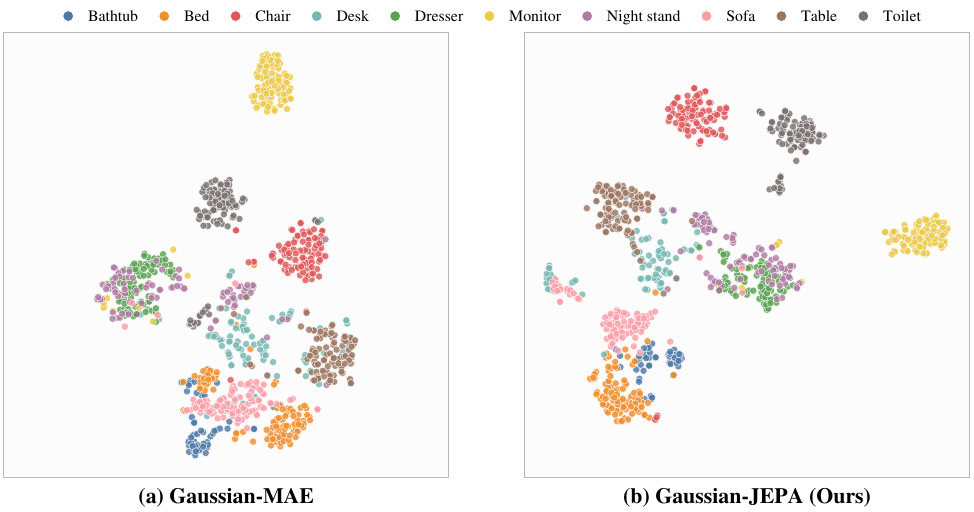}
    \caption{t-SNE visualization of frozen ModelNet10-GS representations.
    Both panels contain the same 908 test objects, colored by category. PCA
    and t-SNE are fitted independently for each method with identical
    hyperparameters; coordinates are therefore not compared across panels.}
    \label{fig:supp:tsne}
\end{figure}

\noindent\textbf{Global embedding organization.}
Fig.~\ref{fig:supp:tsne} compares frozen representations of the same 908 ModelNet10-GS test objects. For each object, we $\ell_2$-normalize three resampled embeddings, average them, and normalize the result. Each method is reduced independently by 50-D PCA followed by t-SNE with perplexity 30 and random seed 0. The panels share objects, labels, colors, and hyperparameters, but their coordinate systems are not aligned and must be interpreted separately. In the original feature space, leave-one-out 10-NN accuracy rises from 87.00\% to 90.09\%, and cosine silhouette from 0.1716 to 0.2425. The visual separation is therefore accompanied by improved neighborhood consistency before dimensionality reduction.

\begin{figure}[t]
    \centering
    \includegraphics[width=\linewidth]{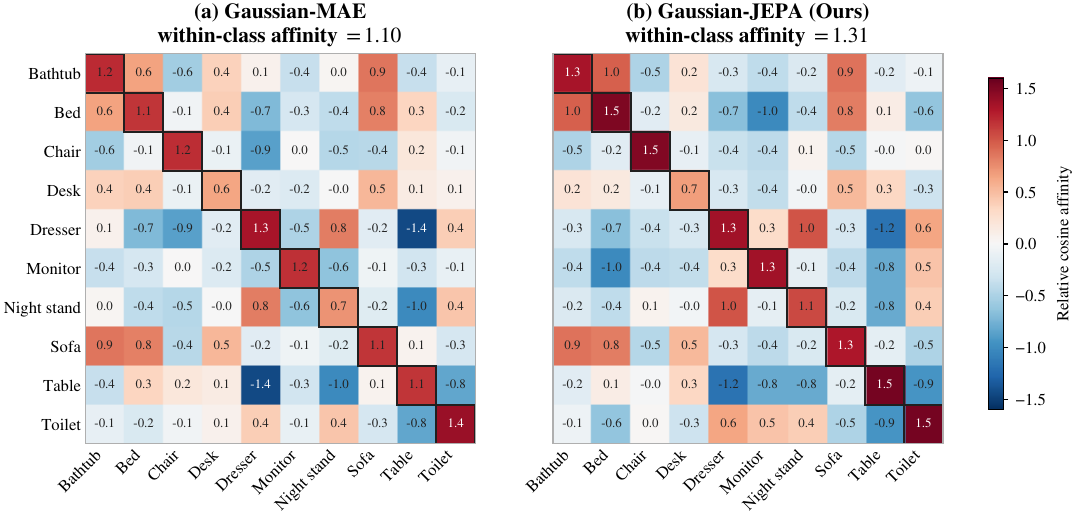}
    \caption{Class-conditional relative cosine affinity in the frozen
    ModelNet10-GS feature spaces. The same 908 objects and three-resampling
    aggregation as Fig.~\ref{fig:supp:tsne} are used. Rows and columns denote
    categories; outlined diagonal cells are within-class affinities. Both
    panels use the same scale.}
    \label{fig:supp:affinity}
\end{figure}

\noindent\textbf{Class-conditional affinity.}
Fig.~\ref{fig:supp:affinity} summarizes pairwise cosine structure before dimensionality reduction. Let $s_{ij}$ denote the cosine similarity between two frozen 768-D representations. For every non-self pair, we compute
\[
a_{ij}=\frac{1}{2}\left(\frac{s_{ij}-\mu_i}{\sigma_i}+\frac{s_{ij}-\mu_j}{\sigma_j}\right),
\]
where $\mu_i$ and $\sigma_i$ are the mean and standard deviation over the non-self gallery for query $i$. We average $a_{ij}$ by class pair. This query-wise normalization removes encoder-specific global cosine offsets; zero denotes the query-specific gallery average, not absolute similarity. \ourmethod{} increases diagonal affinity for all ten categories. The mean rises from 1.10 to 1.31, while the diagonal--off-diagonal margin increases from 1.21 to 1.43. Together with Fig.~\ref{fig:supp:tsne}, this indicates stronger class-relative organization in the original frozen space rather than only in a nonlinear 2-D projection.

\begin{figure*}[t]
    \centering
    \includegraphics[width=\linewidth]{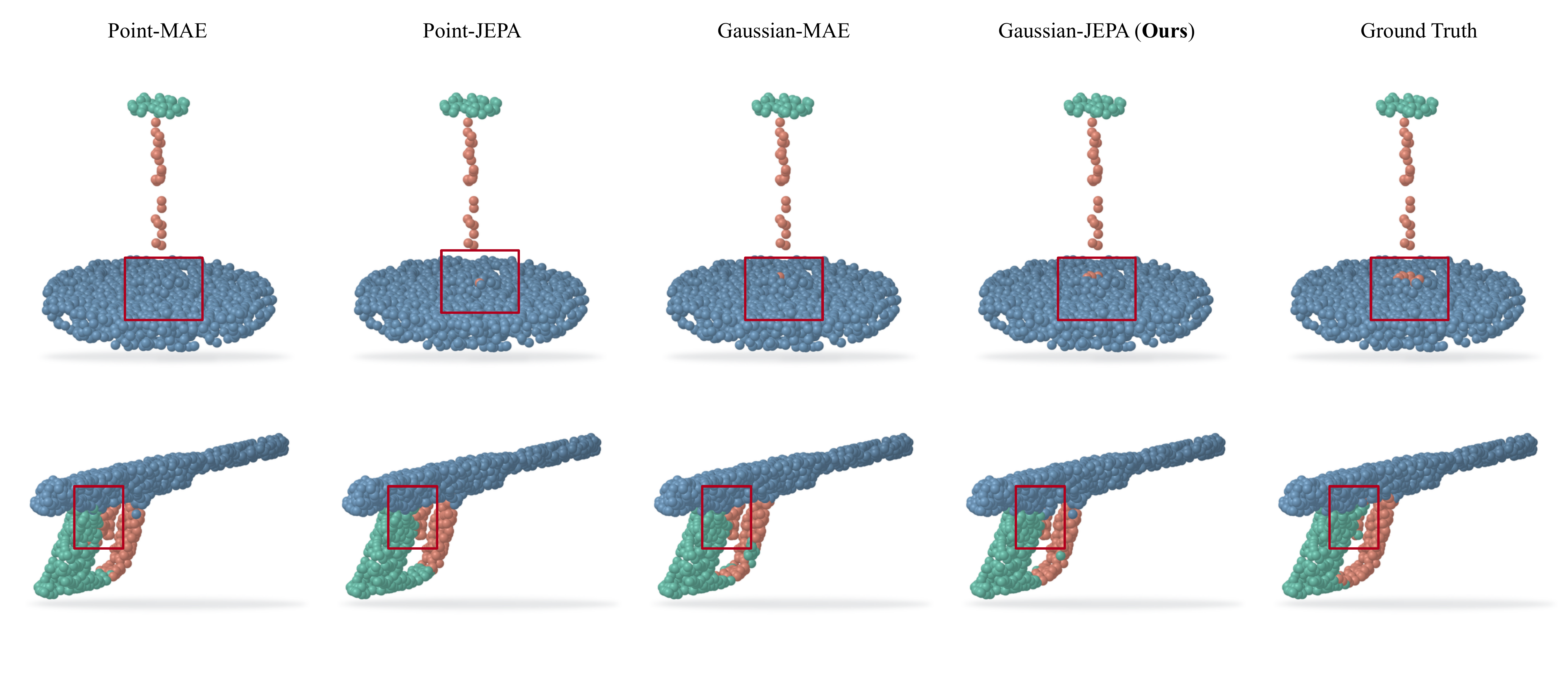}
    \caption{Additional part-segmentation results on ShapeNet-Part.
    Columns compare Point-MAE, Point-JEPA, Gaussian-MAE, \ourmethod{}, and
    ground truth. All methods are visualized on the same 2,048 annotated
    points with a shared camera and part-color mapping.}
    \label{fig:supp:partseg}
\end{figure*}

\noindent\textbf{Part-segmentation predictions.}
Fig.~\ref{fig:supp:partseg} compares representative predictions from point- and Gaussian-pretrained encoders on identical ShapeNet-Part points and cameras. Thin structures and adjacent semantic parts expose boundary errors that aggregate mIoU cannot localize. These examples were not used for model selection.

\begin{figure*}[p]
    \centering
    \includegraphics[width=.83\textwidth]{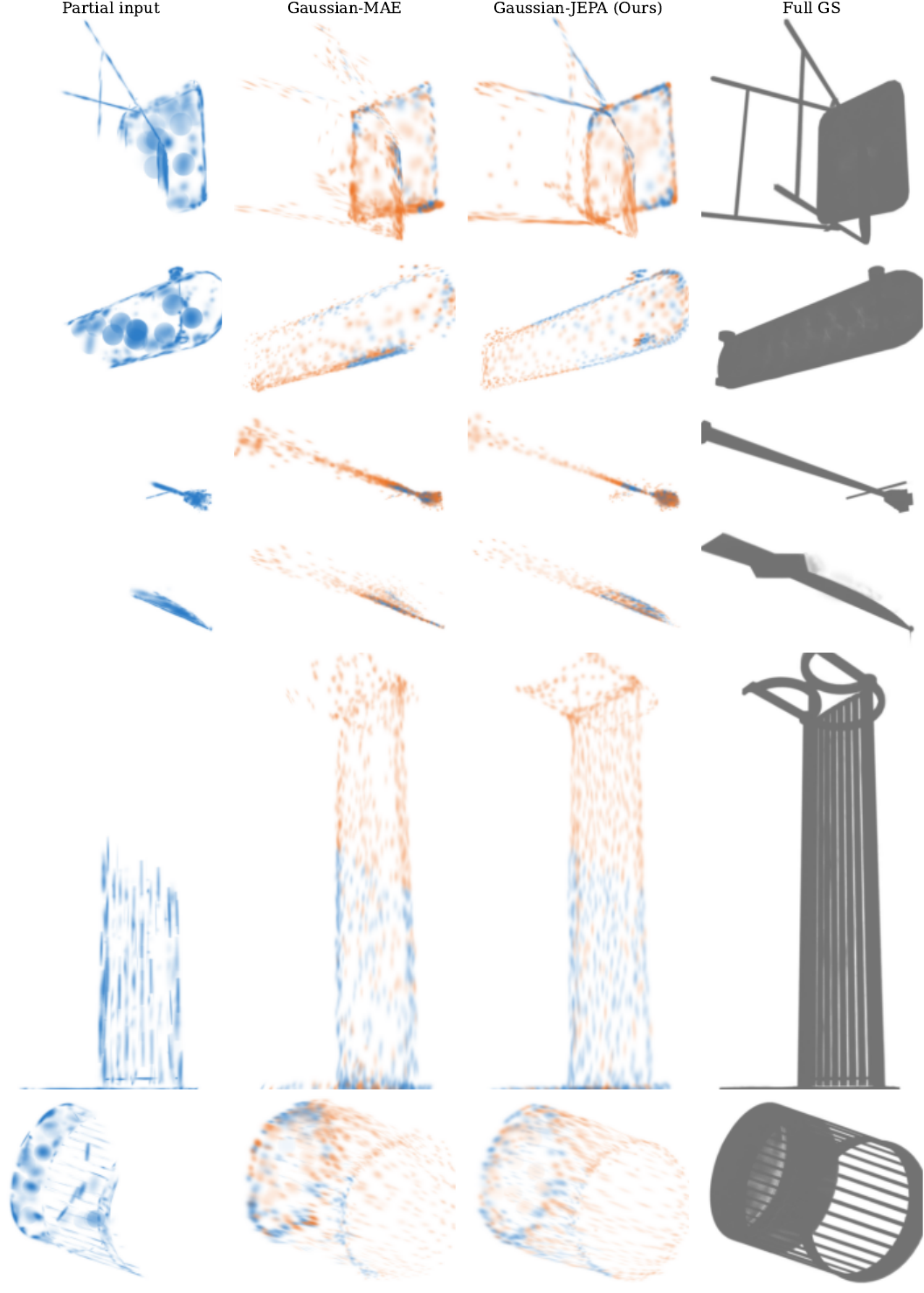}
    \caption{Additional Gaussian shape-completion results on ShapeNet55-GS. From left to right: the partial input (512 GS), the Gaussian-MAE prediction (1K GS), the \ourmethod{} prediction (1K GS), and the full source asset shown as a geometric reference. The same observation, camera viewpoint, and crop are used within each row. Orange highlights the completed regions.}
    \label{fig:supp:completion}
\end{figure*}

\noindent\textbf{Gaussian shape completion.}
Fig.~\ref{fig:supp:completion} provides additional completion examples under the main-paper protocol. Each method receives the same 512-Gaussian partial observation, and identical decoders predict a complete 1K-Gaussian observation from frozen encoder features. The dense source asset is included only as a rendering reference, not as the decoder target. Shared cameras and crops support direct comparison of recovered structure and appearance.

\subsection{Additional Ablations}
The integer target sampler determines the effective mask ratio. Tab.~\ref{tab:supp:mask} compares three hidden-token budgets with four equal-size blocks, isolating mask budget from scale diversity. Relative to 50\%, reducing the mask to 37.5\% raises MN10 linear probing by 0.11 points, while increasing it to 62.5\% lowers accuracy by 0.11 points. The full range spans only 0.22 points. We retain 50\% as a balanced context--target allocation rather than an empirically optimal ratio.

\begin{table}[t]
    \centering
    {\small
    \setlength{\tabcolsep}{5pt}
    \begin{tabularx}{\columnwidth}{@{}>{\raggedright\arraybackslash}Xcc@{}}
    \toprule[0.81pt]
    Target setting & Effective mask & MN10 linear $\uparrow$ \\
    \midrule[0.6pt]
    $[6,6,6,6]$ & $24/64=37.5\%$ & \textbf{93.61} \\
    $[8,8,8,8]$ & $32/64=50.0\%$ & 93.50 \\
    $[10,10,10,10]$ & $40/64=62.5\%$ & 93.39 \\
    \bottomrule[0.81pt]
    \end{tabularx}
    }
    \caption{Target-budget sensitivity. All settings use four equal-size
    targets and feature-space grounding.}
    \label{tab:supp:mask}
\end{table}

\section{Limitations \& Future Work}
Our study focuses on object-level assets under a fixed 1K-Gaussian encoder budget; large scenes, dynamic Gaussians, and adaptive budgets remain open. The resampling evaluation measures stochastic fixed-budget observations of one source asset, whereas equivalence across independently optimized Gaussian assets is a distinct problem. We retain the 14-D input representation and centroid-based grouping of the matched reconstruction baseline. Richer spherical harmonics, anisotropic support, and visibility-aware targets remain promising. Finally, the complementary projections use no geometry or appearance labels; their names denote functional latent views rather than prescribed semantic factors.

\section{Broader Impact}
Learning directly from Gaussian assets may reduce repeated conversion to dense meshes or multi-view images and support reusable 3D perception models. Potential applications include content organization, completion, robotics, and interactive rendering. As with other learned representations, performance depends on the coverage and provenance of pretraining data. Biased, private, or unauthorized assets may propagate harm to downstream systems. Deployments should document data sources, respect ownership and privacy, and evaluate failures under incomplete or out-of-distribution inputs.

\clearpage
\clearpage

\end{document}